\documentclass[10pt,twocolumn,letterpaper]{article}

\usepackage{iccv}
\usepackage{times}
\usepackage{epsfig}
\usepackage{graphicx}
\usepackage{amsmath}
\usepackage{amssymb}
\usepackage{epstopdf}
\usepackage{booktabs}
\usepackage{caption}
\usepackage{subcaption} 

\newcommand{\tabitem}{~~\llap{\textbullet}~~}
\newcommand{\bx}{\mathbf{x}}
\newcommand{\bw}{\mathbf{w}}

\newcommand{\calY}{\mathcal{Y}}
\newcommand{\D}{\mathcal{D}}
\newcommand{\btheta}{\boldsymbol\theta}
\newcommand{\NB}{\text{NB}}

\usepackage[pagebackref=true,breaklinks=true,letterpaper=true,colorlinks,bookmarks=false]{hyperref}

\newcommand{\myparagraph}[1]{\vspace{.5em}\noindent\textbf{#1}}
\newcommand{\ep}[1]{\emph{(#1)}}
\newcommand{\Eq}{Eq.\xspace}
\newcommand{\Fig}{Fig.\xspace}
\newcommand{\Eqs}{Eqs.\xspace}
\newcommand{\Sec}{Sec.\xspace}
\newcommand{\Tab}{Tab.\xspace}

\newcommand{\iid}{\textit{i.i.d.}\xspace}

\iccvfinalcopy % *** Uncomment this line for the final submission

\def\iccvPaperID{1160} % *** Enter the ICCV Paper ID here

\ificcvfinal\fi
\begin{document}

%%%%%%%%% TITLE
\title{DeepSetNet: Predicting Sets with Deep Neural Networks}

\author{S. Hamid Rezatofighi\quad Vijay Kumar B G \quad Anton Milan\quad \\Ehsan Abbasnejad\quad Anthony Dick\quad Ian Reid\\
	School of Computer Science, The University of Adelaide, Australia
	%	{\tt\small hamid.rezatofighi@adelaide.edu.au}
}

\maketitle
%\thispagestyle{empty}

%\thispagestyle{empty}	

%%%%%%%%%%%%%%%%%%%%%%%%%%%%%%%%%%%%%%%%%%%%%%%%%%%%%%%%%%%%%%%%%%%%%%
\begin{abstract}
This paper addresses the task of \emph{set} prediction using deep learning. 
This is important because the output of many computer vision tasks, including image tagging and object detection, are naturally expressed as sets of entities rather than vectors. As opposed to a vector, the size of a set is not fixed in advance, and it is invariant to the ordering of entities within it. 
We define a likelihood for a set distribution and learn its parameters using a deep neural network. We also derive a loss for predicting a discrete distribution corresponding to set cardinality. Set prediction is demonstrated on the problem of multi-class image classification. 
Moreover, we show that the proposed cardinality loss can also trivially be applied to the tasks of object counting and pedestrian detection.
%\fixmeh{It is not clear how set and cardinality is related. Mathematically we are not learning a cardinality distribution only} 
Our approach outperforms existing methods in all three cases on standard datasets.

% The solution to some of these problems can be described by a vector of a known dimension. 
% For example, the goal of semantic segmentation is to label each pixel with a certain category.
% However, problems like image tagging or object detection do no require a specific order in which the tags or bounding boxes should be returned. 
% In other words, the task here is to predict a \emph{set}, \ie an unordered collection of entities of an unknown cardinality. 
% Traditional CNN architectures cannot address this task directly because both the order and the cardinality of the output is firmly built in. 
% In this paper, we propose a method for set prediction using deep learning. 
% To that end, we derive a loss for predicting a discrete distribution corresponding to the set cardinality and show its application on two relevant problems: multi-class image classification and pedestrian detection.
% Our approach yields state-of-the-art results in both cases on standard datasets.
\end{abstract}
\section{Introduction}
\label{sec:introduction}

Deep neural networks have shown state-of-the-art performance on many computer vision problems, including semantic segmentation~\cite{Papandreou:2015:ICCV}, visual tracking~\cite{Nam:2016:CVPR}, image captioning~\cite{Johnson:2016:CVPR}, scene classification~\cite{Krizhevsky:2012:NIPS}, and object detection~\cite{Liu:2016:ECCV}. 
%Despite their remarkable performance, 
However, traditional convolutional architectures require a problem to be formulated in a certain way: in particular, they are designed to predict a \emph{vector} (or a matrix, or a tensor in a more general sense), that is either of a fixed length or whose size depends on the input (\cf fully convolutional architectures). 

\begin{figure}[t]
\centering
\includegraphics[width=1\linewidth]{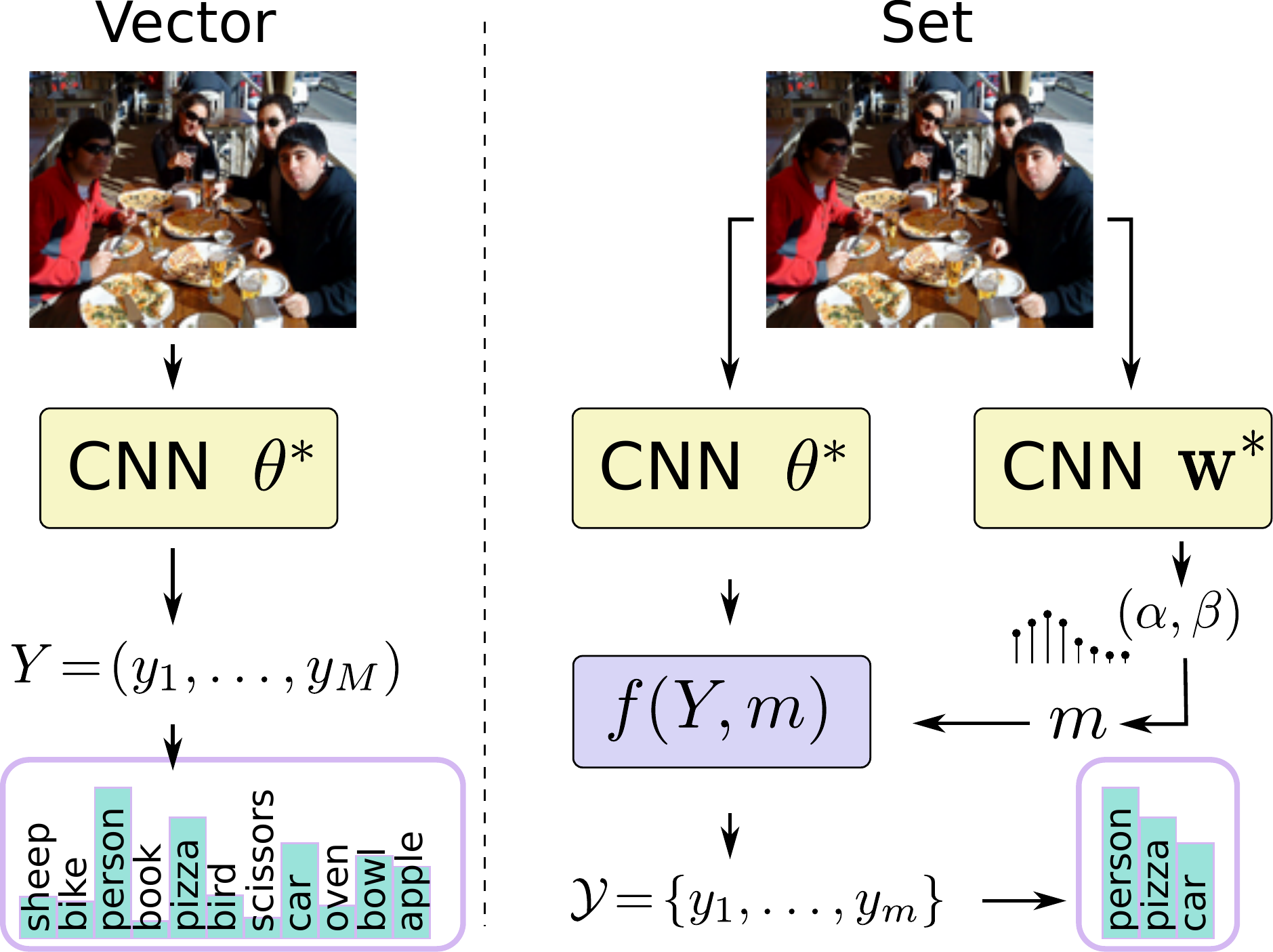}
\caption{{\bf Left:} Traditional CNNs learn parameters $\btheta^*$ to predict a fixed \emph{vector} $Y$. {\bf Right:} In contrast, we propose to train a separate CNN to learn a parameter vector $\bw^*$, which is used to predict the set cardinality of a particular output.}
\vspace{-1em}
\label{fig:method}
\end{figure}

For example, consider the task of scene classification where the goal is to predict the label (or category) of a given image. 
Modern approaches typically address this by a series of convolutional layers, followed by a number of fully connected layers, which are finally mapped to predict a fixed-sized vector~\cite{Krizhevsky:2012:NIPS, Simonyan:2014:VGG, Szegedy:2014:Inception}.
The length of the predicted vector corresponds to the number of candidate categories, \eg 1,000 for the ImageNet challenge~\cite{Russakovsky:2015:ILSVRC}. Each element is a score or probability of a particular category, and the final prediction is a probability distribution over all categories. 
%The mode of that distribution, \ie the entry with the highest score is then taken as the class prediction \fixmeh{I am not sure this is statement is correct. The output of softmax is already the mode of joint distribution of all classes}.
%
This strategy is perfectly admissible if one expects to find exactly one or at least the same number of categories across all images. However, natural images typically show multiple entities (\eg table, pizza, person, \etc), and what is perhaps more important, this number differs from image to image. 
During evaluation, this property is not taken into account. 
The ImageNet Large Scale Visual Recognition Challenge (ILSVRC) only counts an error if the ``true'' label is not among the top-5 candidates. 
Another strategy to account for multiple classes is to fix the number to a certain value for all test instances, and report precision and recall by counting false positive and false negative predictions, as was done in~\cite{gong2013deep,Wang_2016_CVPR}. 
Arguably, both methods are suboptimal because in a real-world scenario, where the correct labelling is unknown, the prediction should in fact not only rank all labels according to their likelihood of being present, but also to report \emph{how many} objects (or labels) are actually present in one particular image.
Deciding how many objects are present in an image is a crucial part of human perception and scene understanding but is missing from our current evaluation of automated image understanding methods.

As a second example, let us consider pedestrian detection. 
The parallel to scene classification that we motivated above is that, once again, in real scenarios, the number of people in a particular scene is not known beforehand. 
The most common approach is to assign a confidence score to a number of region candidates~\cite{Dalal:2005:CVPR, Felzenszwalb:2010:PAMI, Girshick:2015:ICCV, Ren:2015:NIPS}, which are typically selected heuristically by thresholding and non-maxima suppression.
We argue that it is important not to simply discard the information about the actual number of objects at test time, but to exploit it while selecting the subset of region proposals.

The examples above motivate and underline the importance of \emph{set prediction} in certain applications. 
It is important to note that, in contrast to vectors, a set is a collection of elements which is \emph{invariant under permutation} and the size of a set is \emph{not fixed} in advance.
%\fixmeh{The following sentences about thresholding are more confusing sentence to my view. is it ok to remove it???} \fixmea{Sure.} One naive way of extracting a set from a distribution is to apply thresholding and always report all predictions above a fixed confidence level.
%One single global threshold, however, cannot produce the desired effect because it is highly dependent on each data sample and should therefore be adjusted for each instance.
%However, an adaptive threshold per training instance cannot be learned in a straightforward manner as it is not directly observable from the training data. 
To this end, we use a principled definition of a set as the union of cardinality distribution and family of joint distributions over each cardinality value.
In summary, our main contributions are as follows:
\emph{(i)}
Starting from the mathematical definition of a set distribution, we derive a loss that enables us to employ existing machine learning methodology to learn this distribution from data.
\emph{(ii)}
We integrate our loss into a deep learning framework to exploit the power of a multi-layer architecture.
\emph{(iii)}
 We present state-of-the-art results for multi-label image classification and pedestrian detection on standard datasets and competitive results on the task of object counting.
% \begin{itemize}
% \setlength\itemsep{.1em}
% \item Starting from the mathematical definition of a set distribution, we derive a loss that enables us to employ existing machine learning methodology to learn this distribution from data.
% \item We integrate our loss into a deep learning framework to exploit the power of a multi-layer architecture.
% \item We present state-of-the-art results on standard datasets on the tasks of object counting, pedestrian detection and multi-label image classification.
% \end{itemize}

% A somewhat different approach is to use fully convolutional networks~\cite{Long:2015:CVPR, Chen:2014:ICLR, Chen:2016:arxiv} that replace 

% * Deep Learning is successful in many applications (image classification, semantic labeling, pedestrian detections, etc.)
% 
% * Many settings require a varying number of solutions (e.g. image tags), i.e. they are inherently defined to output a set.
% 
% * Moreover, the ordering of the output has no semantic meaning (see set).
% 
% * Typical CNNs require a fixed output size (e.g. number of classes)
% 
% * It is important to preserve the (distribution) of cardinality from the input.
% 
% * Most CNNs learn feature VECTOR to feature VECTOR mapping. I.e. it is not permutation invariant.
% 
% * We need to include cardinality.
% 
% * RNNs can be used to predict variable length outputs (e.g. machine translation, image captioning)
% 
% * RNNs are not suitable here because they rely on a fixed ordering. 

\section{Related Work}
% \myparagraph{Related Work.}
\label{sec:relwork}
A sudden success in multiple applications including voice recognition~\cite{Graves:2013:ICASSP}, machine translation~\cite{Sutskever:2014:NIPS} and image classification~\cite{Krizhevsky:2012:NIPS}, has sparked the deployment of deep learning methods throughout numerous research areas. 
Deep convolutional (CNN) and recurrent (RNN) neural networks now outperform traditional approaches in tasks like semantic segmentation~\cite{Chen:2014:ICLR, Lin:2017:CVPR}, image captioning~\cite{Johnson:2016:CVPR} or object detection~\cite{Liu:2016:ECCV}.
Here, we will briefly review some of the recent approaches to image classification and object detections and point out their limitations.

Image or scene classification is a fundamental task of understanding photographs.
The goal here is to predict a scene label for a given image. 
Early datasets, such as \mbox{Caltech-101}~\cite{FeiFei:2006:PAMI}, mostly contained one single object and could easily be described by one category. Consequently, a large body of literature focused on single-class prediction~\cite{Krizhevsky:2012:NIPS, Sermanet:2013:OverFeat, Zeiler:2014:ECCV, Murthy:2016:CVPR}.
However, real-world photographs typically contain a collection of multiple objects and should therefore be captioned with multiple tags. 
% Each image in the latest edition of the ImageNet Large Scale Visual Recognition Challenge (ILSVRC)~\cite{Russakovsky:2015:ILSVRC} is labelled with multiple classes. The performance of an algorithm is measured as precision and recall \fixmea{check!}
%To that end, 
%
Surprisingly, there exists rather little work on multi-class image classification that makes use of deep architectures. Gong~\etal~\cite{Gong:2013:arxiv} combine deep CNNs with a top-$k$ approximate ranking loss to predict multiple labels. Wei~\etal~\cite{Wei:2014:arxiv} propose a Hypotheses-Pooling architecture that is specifically designed to handle multi-label output. While both methods open a promising direction, their underlying architectures largely ignore the correlation between multiple labels.
To address this limitation, recently, Wang~\etal~\cite{Wang_2016_CVPR} combined CNNs and RNNs to predict a number of classes in a sequential manner. 
RNNs, however, are not suitable for set prediction mainly for two reasons. 
First, the output represents a sequence rather than a set and is thus highly dependent on the prediction order, as was shown recently by Vinyals~\etal~\cite{Vinyals:2015:arxiv}. 
Second, the final prediction may not result in a feasible solution (\eg it may contain the same element multiple times), such that post-processing or heuristics such as beam search must be employed~\cite{Vinyals:2015:NIPS, Wang_2016_CVPR}. 
Here we show that our approach not only guarantees to always predict a valid set, but also outperforms previous methods. 

%However, they need heuristics like beam search to guarantee that the predicted sequence does not include repetitions. In contrast, our proposed deep network is specifically designed for the task of predicting the cardinality of the output~\fixmea{mathematically derived...}

%Recurrent neural networks (RNNs) present a viable alternative to CNNs as they are designed to predict sequences of potentially arbitrary length. 
%A CNN-RNN approach has indeed recently been investigated for the task of multi-label image classification~\cite{Wang_2016_CVPR}. 

Pedestrian detection can also be viewed as a classification problem. Traditional approaches follow the sliding-window paradigm~\cite{Viola:2004:IJCV, Dalal:2005:CVPR, Walk:2010:CVPR, Felzenszwalb:2010:PAMI, Benenson:2012:CVPR}, where each possible (or rather plausible) image region is scored independently to contain a person or not. 
% Prominent examples of these techniques include the Viola and Jones detector that uses local binary patterns as features~\cite{Viola:2004:IJCV}, the Histogram-of-Gradients features of Dalal and Triggs~\cite{Dalal:2005:CVPR}, or the deformable part-based model (DPM) that allows for local deformation of body parts to better capture the non-rigid property of certain object categories.
More recent methods like Fast R-CNN~\cite{Girshick:2015:ICCV} or the single-shot multi-box detector (SSD)~\cite{Liu:2016:ECCV} learn the relevant image features rather than manually engineering them, but retain the sliding window approach.

All the above approaches require some form of post-processing to suppress spurious detection responses that originate from the same person. This is typically addressed by non-maximum suppression (NMS), a greedy optimisation strategy with a fixed overlap threshold. 
Recently, several alternatives have been proposed to replace the greedy NMS procedure, including sequential head detection with LSTMs~\cite{Hochreiter:1997:LSTM}, a global optimisation approach to NMS~\cite{Pham:2016:CVPR,Lee:2016:ECCV}, or even learning NMS end-to-end using CNNs~\cite{Hosang:2017:CVPR}. None of the above methods, however, explicitly consider the number of objects while selecting the final set of boxes. Contrary to existing pedestrian detection approaches, we incorporate the cardinality into the NMS algorithm itself and show an improvement over the state of the art on two benchmarks. Note that the idea of considering the number of pedestrians can be applied in combination with any of the aforementioned detection techniques to further improve their performances.

It is important to bear in mind that unlike many existing approaches that learn to count~\cite{arteta2014interactive,chan2009bayesian,fiaschi2012learning,idrees2013multi,lempitsky2010learning, pham2015count, zhang2015cross,zhang2016single}, our main goal is not object counting. Rather, we derive a formulation for set prediction using deep learning. Estimating the cardinality of objects and thereby their count is a byproduct of our approach. To demonstrate the effectiveness of our formulation, we also conduct experiments on the task of object counting, outperforming many recent methods on the widely used USCD dataset.

% Russel~\etal~\cite{Stewart:2016:CVPR} perform end-to-end head detection by predicting the bounding boxes sequentially using an LSTM~\cite{Hochreiter:1997:LSTM}. Their approach, however, processes only a small image region at a time, thus limiting the applicability on crowded scenarios with tens or hundreds of people. 
% %
% Pham~\etal~\cite{Pham:2016:CVPR} and Lee~\etal~\cite{Lee:2016:ECCV} formulate NMS as a global optimisation problem
% while Hosang~\etal~\cite{Hosang:2016:GCPR} propose to learn the NMS algorithm end-to-end using CNNs. Both methods, however, do not consider the number of objects while selecting the final set of boxes. Contrary to existing pedestrian detection approaches, we incorporate the cardinality into the NMS algorithm itself. This leads to an improvement over the state of the art, validated on two benchmarks.
\section{Random Vectors vs. Random Finite Sets} 
\label{sec:background}
% To explain our approach, we first review some mathematical background and introduce the notation used throughout the paper.

In statistics, a continuous random variable $y$ is a variable that can take an infinite number of possible values. A continuous random vector can be defined by stacking several continuous random variables into a fixed length vector, $Y=\left(y_1,\cdots,y_m\right)$. The mathematical function describing the possible values of a continuous random vector and their associated joint probabilities is known as a probability density function (PDF) $p(Y)$ such that
$\int p(Y)dY = 1.$
%**********************************************************************************************
% \begin{equation}\label{eq: pdf}
% \begin{aligned}
% %\int_{-\infty}^{\infty} p(Y)dY = 1.
% \int p(Y)dY = 1.
% \end{aligned}
% \end{equation}
%**********************************************************************************************

In contrast, a random finite set (RFS) $\calY$ is a finite-set valued random variable $\calY=\left\{y_1,\cdots,y_m\right\}$. The main difference between an RFS and a random vector is that for the former, the number of constituent variables, $m$, is random and the variables themselves are random and unordered. Throughout the paper, we use $\calY=\left\{y_1,\cdots,y_m\right\}$ for a set with unknown cardinality, $\calY_m=\left\{y_1,\cdots,y_m\right\}_{||}$ for a set with known cardinality and $Y=\left(y_1,\cdots,y_m\right)$ for a vector with known dimension. 
%\fixmea{Somehow, I find it a bit confusing to call both known and unknown number of elements with the same variable $m$. Also, is $_{||}$ some standard notation?}  

A statistical function describing a finite-set variable $\calY$ is a
combinatorial probability density function $p(\calY)$ which consists of a discrete probability distribution, the so-called cardinality distribution, and a family of joint probability densities on both the number and the values of the constituent variables. 
Similar to the definition of a PDF for a random variable, the PDF of an RFS must sum to unity over all possible cardinality values and all possible element values and their permutations~\cite{mahler2007statistical}.
%, \ie
%%**********************************************************************************************
%\begin{equation}\label{eq: RFS pdf}
%\begin{aligned}
%\int p(\calY)&\delta \calY \triangleq p(\emptyset) +\\
%&\sum_{m=1}^{\infty}\frac{1}{m!}\int p(\{y_1,\cdots,y_m\}_{||}) dy_1\cdots dy_m = 1,
%\end{aligned}
%\end{equation}
%%**********************************************************************************************
%where %$\int \cdot\delta \calY$ is integral set and calculated as shown above and
% $p(\emptyset)$ is the probability of the empty set. 
The PDF of an $m$-dimensional random vector can be defined in terms of an RFS as
%**********************************************************************************************
\begin{equation}\label{eq: pdf rfs vs vect}
\begin{aligned}
p(y_1,\cdots,y_m) \triangleq \frac{1}{m!} p(\{y_1,\cdots,y_m\}_{||}).
\end{aligned}
\end{equation}
%**********************************************************************************************
The normalisation factor $m!=\prod_{k=1}^m k$ appears because the probability density for a set $\{y_1,\cdots,y_m\}_{||}$ must be equally distributed among all the $m!$ possible permutations of the vector~\cite{mahler2007statistical}.

Conventional machine learning approaches, such as Bayesian learning and convolutional neural networks, have been proposed to learn the optimal parameters $\btheta^*$ of the distribution $p(Y|\bx,\btheta^*)$ which maps the input vector $\bx$ to the \emph{output vector} $Y$.
In this paper, we instead propose an approach that can learn a pair $(\btheta^*,\bw^*)$ of parameter vectors for a set distribution that allow one to map the input vector $\bx$ into the \emph{output set} $\calY$, i.e.  $p(\calY|\bx, \btheta^*,\bw^*)$. The additional parameters $\bw^*$ define a PDF over the set cardinality, as we will explain in the next section.
\section{Deep Set Network}
\label{sec:deep-set-net}

Let us begin by defining a training set $\D = \{\calY_{i},\bx_{i}\}$,
% \begin{eqnarray*}
% 	\calY & = & \{\by_{1},\by_{2},\ldots,\by_{m}\}\qquad \by_{k}\in\mathbb{R}^{d}, \forall k\\ % \text{or}\quad\mathbb{Z}^{d} \\
% 	\D & = & \{\calY_{i},\bx_{i}\}\qquad\qquad\quad\quad\forall\bx_{i}\in\mathbb{R}^{l}\quad\forall i=1,\ldots,n,
% \end{eqnarray*}
where each training sample $i=1,\ldots,n$ is a pair consisting of an input feature $\bx_{i}\in\mathbb{R}^{l}$ and an output (or label) set
$\calY_{i} = \{y_{1},y_{2},\ldots,y_{m_i}\}, y_{k}\in\mathbb{R}^{d} $, $m_i\in\mathbb{N}^0$. In the following we will drop the instance index $i$ for better readability. Note that $m:=|\calY|$ denotes the cardinality of set $\calY$.
The probability of a set $\calY$ is defined as
\begin{equation}
\begin{aligned}
	p(\calY|\bx,\btheta,\bw) = & p(m|\bx,\bw)\times\\
	& m!\times U^m\times p(y_{1},y_{2},\cdots,y_{m}|\bx,\btheta),
	\label{eq:set_density}
\end{aligned}   
\end{equation} 
where $p(m|\cdot,\cdot)$ and $ p(y_{1},y_{2},\cdots,y_{m}|\cdot,\cdot)$ are respectively a cardinality distribution and a
symmetric joint probability density distribution of the elements. $U$ is the unit of hypervolume
in $\mathbb{R}^{d}$, which makes the joint distribution unitless~\cite{mahler2007statistical,vo2016model}. $\btheta$ denotes the parameters that estimate the joint distribution of set element values for a fixed cardinality,\footnote{This is also known as \emph{spatial distribution of points} in point process statistics.}
while $\bw$ represents the collection of parameters which estimate the cardinality distribution of the set elements.

The above formulation represents the probability density of a set which is very general and completely independent from the choices of both cardinality and spatial distributions. It is thus straightforward to transfer it to many applications that require the output to be a set.  However, to make the problem amenable to mathematical derivation and implementation, we adopt two assumptions: \emph{i)} the outputs (or labels) in the set are independent
and identically distributed (\iid) and  \emph{ii)} their cardinality follows
a Poisson distribution with parameter $\lambda$. Thus, we can write the distribution as
\begin{equation}
\begin{aligned}
	p(\calY|\bx,\btheta,\bw) = \int p(m|\lambda)&p(\lambda|\bx,\bw) d\lambda\times\\
  m!\times U^m\times&\left(\prod_{k=1}^{m}p(y_{k}|\bx,\btheta)\right).
\end{aligned}   
\end{equation}

\subsection{Posterior distribution}
\label{sec:posterior}
To learn the parameters $\btheta$ and $\bw$, we first define the posterior distribution over them as
\begin{equation}
\begin{aligned}
	p(\btheta,\bw|\D) &\propto p(\D|\btheta,\bw)p(\btheta)p(\bw)\\
	&\propto\prod_{i=1}^{n}\left[\int p(m_{i}|\lambda)p(\lambda|\bx_{i},\bw)d\lambda\times m_{i}! \right.\\ &\left.U^{m_{i}}\left(\prod_{k=1}^{m_{i}}p(y_{k}|\bx_{i},\btheta)\right)\right]p(\bx_{i})p(\btheta)p(\bw).
\end{aligned} 
\label{eq:posterior0}
\end{equation}
A closed form solution for the integral in \Eq~\eqref{eq:posterior0} can be obtained by using conjugate priors:
\begin{eqnarray*}
	m & \sim & \mathcal{P}(m;\lambda)\\
	\lambda & \sim & \mathcal{G}(\lambda;\alpha(\bx,\bw),\beta(\bx,\bw))\\
	&&\alpha(\bx,\bw),\beta(\bx,\bw)>0\quad\forall\bx,\bw\\
	\btheta & \sim & \mathcal{N}(\btheta;0,\sigma_{1}^{2}\mathbf{I})\\
	\bw & \sim & \mathcal{N}(\bw;0,\sigma_{2}^{2}\mathbf{I}),
\end{eqnarray*}
where $\mathcal{P}(\cdot,\lambda)$, $\mathcal{G}(\cdot;\alpha,\beta)$,  and $\mathcal{N}(\cdot;0,\sigma^{2}\mathbf{I})$ represent respectively a Poisson distribution with parameters $\lambda$, a Gamma distribution with parameters $(\alpha,\beta)$ and a zero mean normal distribution with covariance equal to $\sigma^{2}\mathbf{I}$. 

We assume that the cardinality follows a Poisson distribution whose mean, $\lambda$, follows a Gamma distribution, with parameters which can be estimated from the input data $\bx$. 
Note that the cardinality distribution in \Eq~\eqref{eq:set_density} can be replaced by any other discrete distribution. For example, it is a valid assumption to model the number of objects in natural images by a Poisson distribution~\cite{chan2009bayesian}. Thus, we could directly predict $\lambda$ to model this distribution by formulating the cardinality as $p(m|\bx,\bw) = \mathcal{P}(m;\lambda(\bx,\bw))$. However, this would limit the model's expressive power because two visually entirely different images with the same number of objects would be mapped to the same $\lambda$. Instead, to allow for uncertainty of the mean, we model it with another distribution, which we choose to be Gamma for mathematical convenience.  
Consequently, the integrals in $p(\btheta,\bw|\D)$ are simplified
and form a negative binomial distribution,
\begin{equation}
	\NB\left(m;a,b\right)  = \frac{\Gamma(m+a)}{\Gamma(m+1)\Gamma(a)}.(1-b)^{a}b^{m}, 
\end{equation}
where $\Gamma$ is the Gamma function. Finally, the full posterior distribution can be written as
\begin{equation}
\begin{aligned}
p(\btheta,\bw|\D) & \propto\prod_{i=1}^{n}\bigg[\NB\left(m_{i};\alpha(\bx_{i},\bw),\frac{1}{1+\beta(\bx_{i},\bw)}\right)\\
&\!\!\!\!\!\!\!\!\times m_{i}!\times U^{m_{i}}\times\left(\prod_{k=1}^{m_{i}}p(y_{k}|\bx_{i},\btheta)\right)\bigg]p(\btheta)p(\bw).
\label{eq:full-posterior0}
\end{aligned}   
\end{equation}

\subsection{Learning}
\label{sec:learning}

For simplicity, we use a point estimate for the posterior $p(\btheta,\bw|\D)$,
\ie $p(\btheta,\bw|\D) = \delta(\btheta=\btheta^{*},\bw=\bw^{*}|\D)$, where $(\btheta^{*},\bw^{*})$ are computed using the following MAP estimator: 
\begin{equation}
	(\btheta^{*},\bw^{*}) = \arg\max_{\btheta,\bw}\quad \log\left(p\left(\btheta,\bw|\D\right)\right).
	\label{eq:map0}
\end{equation}
The optimisation problem in \Eq~\eqref{eq:map0} can be decomposed \wrt the parameters
$\btheta$ and $\bw$. Therefore, we can learn them independently as
\vspace{-.5em}
\begin{equation}
	\btheta^{*} =  \arg\max_{\btheta}\quad-\gamma_{1}\|\btheta\|+\sum_{i=1}^{n}\sum_{k=1}^{m_{i}}\log\left(p(y_{k}|\bx_{i},\btheta)\right)
	\label{eq:CNN_Eq}
\end{equation}
and 
\begin{equation}
\begin{aligned}
	\bw^{*} 
	= & \arg\max_{\bw}\quad\sum_{i=1}^{n}\Big[\log\left(\frac{\Gamma(m_{i}+\alpha(\bx_{i},\bw))}{\Gamma(m_{i}+1)\Gamma(\alpha(\bx_{i},\bw))}\right)\\
	+ & \displaystyle{ \log\left(\frac{\beta(\bx_{i},\bw)^{\alpha(\bx_{i},\bw)}}{(1+\beta(\bx_{i},\bw)^{\alpha(\bx_{i},\bw)+m_{i}})}\right)}\Big]-\gamma_2\|\bw\|,
	\label{eq:Cardinal_Eq0}
\end{aligned}   
\end{equation}
where $\gamma_1$ and $\gamma_2$ are the regularisation parameters, proportional to the predefined covariance parameters $\sigma_1$ and $\sigma_2$. These parameters are also known as weight decay parameters and commonly used in training neural networks.
%\fixmea{Check:} It is important to note that the independence of $\btheta^*$ and $\bw^*$ is not due to some simplifying assumption but rather a direct consequence of our formulation for set prediction.

The learned parameters $\btheta^{*}$ in \Eq~\eqref{eq:CNN_Eq} are used to map an input feature vector $\bx$  into an output vector $Y$. For example, in image classification, $\btheta^*$ is used to predict the distribution $Y$ over all categories, given the input image $\bx$. Note that $\btheta^*$ can generally be learned using a number of existing machine learning techniques. In this paper we rely on deep CNNs to perform this task.

To learn the highly complex function between the input feature $\bx$ and the parameters $(\alpha,\beta)$, which are used for estimating the output cardinality distribution, we train a second deep neural network. 
Using neural networks to predict a discrete value may seem counterintuitive, because these methods at their core rely on the backpropagation algorithm, which assumes a differentiable loss. Note that we achieve this by describing the discrete distribution by continuous parameters $\alpha, \beta$ (Negative binomial $\NB(\cdot,\alpha,\frac{1}{1+\beta})$), and can then easily draw discrete samples from that distribution. More formally, to estimate $\bw^{*}$, we compute the partial derivatives of the objective function in \Eq~\eqref{eq:Cardinal_Eq0} \wrt $\alpha (\cdot,\cdot)$ and $\beta (\cdot,\cdot)$ and use standard backpropagation to learn the parameters of the deep neural network.

We refer the reader to the supplementary material for the complete derivation of the partial derivatives, a more detailed derivation of the posterior in \Eqs~\eqref{eq:posterior0}-\eqref{eq:full-posterior0} and the proof for decomposition of the MAP estimation in \Eq~\eqref{eq:map0}.

\subsection{Inference}
\label{sec:inference}
Having the learned parameters of the network $(\bw^{*},\btheta^{*})$, for a test feature $\bx^{+}$, we use a MAP estimate to generate a set output as 
% \begin{equation}
$
\calY^{*} 
= \arg\max_{\calY} p(\calY|\D,\bx^{+}),
$
% \end{equation}
where
\begin{eqnarray*}
	p(\calY|\D,\bx^{+}) & = & \int p(\calY|\bx^{+},\btheta,\bw)p(\btheta,\bw|\D)d\btheta d\bw
\end{eqnarray*}
and $p(\btheta,\bw|\D) = \delta(\btheta=\btheta^{*},\bw=\bw^{*}|\D)$. 
Since the unit of hypervolume $U$ in most practical application in unknown, to calculate the mode of the set distribution $p(\calY|\D,\bx^{+})$, we use sequential inference as explained in~\cite{mahler2007statistical}. To this end, we first calculate the mode $m^*$ of the cardinality distribution 
$
% \begin{equation}
m^{*} 
= \arg\max_{m} p(m|\bx^{+},\bw^*),
% \label{eq:mode_card}
% \end{equation}
$
where $p(m|\bx^{+},\bw^*)=\NB\left(m;\alpha(\bx^{+},\bw^*),\frac{1}{1+\beta(\bx^{+},\bw^*)}\right)$.
Then, we calculate the mode of the joint distribution for the given cardinality $m^{*}$ as
\begin{equation}
\calY^{*}
= \arg\max_{\calY_{m^{*}}}\quad p(\{y_1,\cdots,y_{m^{*}}\}_{||}|\bx^{+},\btheta^*).
\end{equation}
    To estimate the most likely set $\calY^{*}$ with cardinality $m^{*}$, we use the first CNN with the parameters $\btheta^*$ which predicts $p(y_1,\cdots,y_{M}|\bx^{+},\btheta^*)$, where $M$ is the maximum cardinality of the set, \ie $\{y_1,\cdots,y_{m^{*}}\}\subseteq\{y_1,\cdots,y_{M}\},\quad\forall m^{*}$.
Since the samples are \iid, the joint probability maximised when the probability of each element in the set is maximised. Therefore, the most likely set $\calY^{*}$ with cardinality $m^{*}$ is obtained by ordering the probabilities of the set elements $y_1,\cdots,y_{M}$ as the output of the first CNN and choosing $m^{*}$ elements with highest probability values. 
 
% \fixmeh{We can say that this can be simply seen as the combination of cardinality estimation and any detector/classifier???}
Note that the assumptions in \Sec~\ref{sec:deep-set-net} are necessary to make both learning and inference computationally tractable and amenable to an elegant mathematical formulation.
A major advantage of this approach is that we can use any state-of-the-art classifier/detector as the first CNN ($\btheta^*$) to further improve its performance. 
Modifying any of the assumptions, \eg non-\iid set elements, leads to serious mathematical complexities~\cite{mahler2007statistical}, and is left for future work.

\section{Experimental Results}
\label{sec:results}

% \fixmea{Now that I'm reading the experiments section, I'm wondering if it would be better to put multi-label classification in front, after all. This is like our main argument, which should be most prominent. We can argue both ways, of course, but the way it is now, it is somehow lost after the two sections 5.1 and 5.2. What do you think?}
% \fixmel{I agree, the multi-label should go first.}

Our proposed method is best suited for applications that expect the solution to be in the form of a set, \ie permutation invariant and of an unknown cardinality. To that end, we perform an
experiment on multi-label image classification in \Sec~\ref{sec:multi-label}. In addition, we explore
our cardinality estimation loss on the object counting problem in \Sec~\ref{sec:crowd-counting} and then
show  in \Sec~\ref{sec:detection} how incorporating cardinality into a state-of-the art pedestrian detector and formulating it as a set problem can boost up its performance.

\subsection{Multi-Label Image Classification}
\label{sec:multi-label}
\begin{figure*}[t]
	\centering
	\includegraphics[width=1\linewidth]{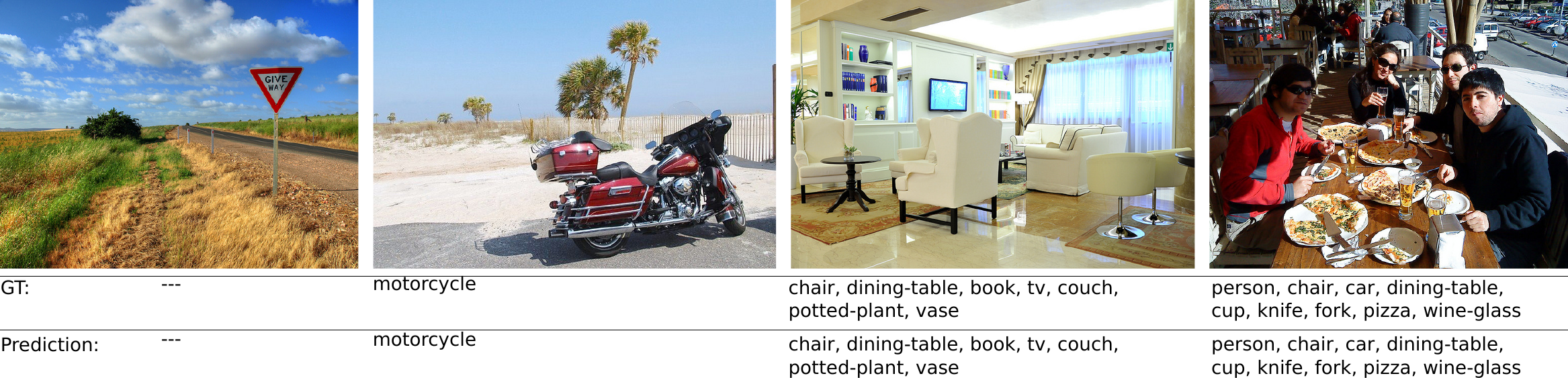}
	\caption{Qualitative results of our multi-class image labelling approach. For each image, the ground truth tags and our predictions are denoted below. Note that we show the exact output of our network, without any heuristics or post-processing.} 
	\label{fig:Results1}
\end{figure*}
\begin{figure*}[t]
	\centering
	\includegraphics[width=1\linewidth]{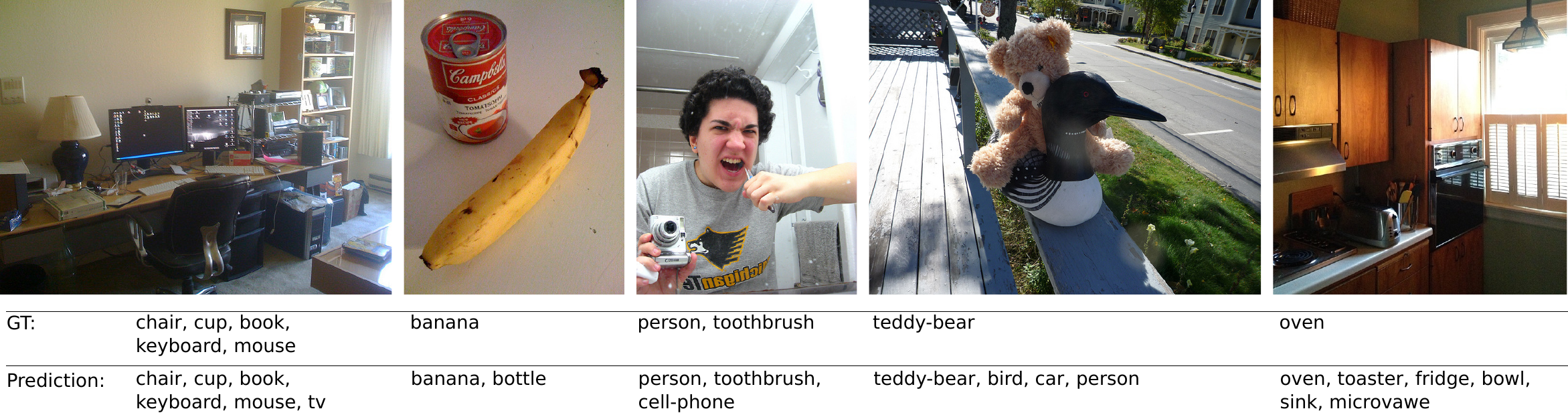}
	\caption{Interesting failure cases of our method. The ``spurious'' TV class predicted on the left is an artifact in annotation because in many examples, computer monitors are actually labelled as TV. In other cases, our network can correctly reason about the number of objects or concepts in the scene, but is constrained by a fixed list of categories defined in the dataset.} 
	\label{fig:Results2}
\end{figure*}
% To demonstrate the effectiveness of our approach, we test it on a highly relevant task of multi-label image classification. 
As opposed to the more common and more studied problem of (single-label) image classification, the task here is rather to label a photograph with an arbitrary, a-priori unknown number of tags. We perform experiments on two standard benchmarks, the PASCAL VOC 2007 dataset~\cite{Everingham:2007:PASCAL-VOC} and the Microsoft Common Objects in Context (MS COCO) dataset~\cite{Lin:2014:COCO}.

\myparagraph{Implementation details.}
% \subsubsection{Implementation details}
In this experiment, similar to~\cite{Wang_2016_CVPR}, we build on the $16$-layers VGG network~\cite{Simonyan:2014:VGG}, pretrained on the 2012 ImageNet dataset. We adapt VGG for our purpose by modifying the last fully connected prediction layer to predict $20$ classes for PASCAL VOC, and $80$ classes for MS COCO. We then fine-tune the entire network for each of these datasets using two commonly used losses for multi-label classification, \textit{softmax} and \textit{binary cross-entropy (BCE)}\footnote{\textit{Weighted Approximate Ranking} (WARP) objective is another commonly used loss for multi-label classification. However, it does not perform as well as \textit{softmax} and \textit{binary cross-entropy} for the used datasets~\cite{Wang_2016_CVPR}.}~\cite{gong2013deep,Wang_2016_CVPR}. To learn both classifiers, we set the weight decay to $5\cdot 10^{-4}$, with a momentum of $0.9$ and a dropout rate of $0.5$.  The learning rate is adjusted to gradually decrease after each epoch, starting from $0.01$ for \textit{softmax} and from $0.001$ for \textit{binary cross-entropy}. The learned parameters of these classifiers correspond to $\btheta^*$ for our proposed deep set network (\cf~\Eq~\eqref{eq:CNN_Eq} and \Fig~\ref{fig:method}).

To learn the cardinality distribution, we use the same VGG-16 network as above and modify the final fully connected layer to predict $2$ values followed by two weighted sigmoid activation functions for $\alpha$ and $\beta$. It is important to note, that the outputs must be positive to describe a valid Gamma distribution. We therefore also append two weighted sigmoid transfer functions 
with weights $\alpha_{M}, \beta_{M}$
% ($\alpha_{M}\sigma(\cdot)$ and $\beta_{M}\sigma(\cdot)$ ) 
to ensure that the values predicted for $\alpha$ and $\beta$ are in a valid range. Our model is not sensitive to these parameters and we set their values to be large enough ($\alpha_{M}=160$ and $\beta_{M}=20$) to guarantee that the mode of the distribution can accommodate the largest cardinality existing in the dataset. We then fine-tune the network on cardinality distribution using the objective loss defined in \Eq~\eqref{eq:Cardinal_Eq0}. To train the cardinality CNN, we set a constant learning rate $0.001$, weight decay $5\cdot10^{-12}$, momentum rate $0.9$ and dropout $0.5$.  

\myparagraph{Evaluation protocol.}
% \subsubsection{Evaluation protocol}
To evaluate the performance of the classifiers and our deep set network, we employ the commonly used evaluation metrics for multi-label image classification~\cite{gong2013deep,Wang_2016_CVPR}: \textit{precision} and \textit{recall} of the generated labels per-class (C-P and C-R) and overall (O-P and O-R). Precision is defined as the ratio of
correctly predicted labels and total predicted labels, while
recall is the ratio of correctly predicted labels and ground-truth labels. 
In case no predictions (or ground truth) labels exist, \ie the denominator becomes zero, precision (or recall) is defined as $100\%$. To generate the predicted labels for a particular image, we perform a forward pass of the CNN and choose top-$k$ labels according to their scores similar to~\cite{gong2013deep,Wang_2016_CVPR}.
Since the classifier always predicts a fixed-sized prediction for all categories, we sweep $k$ from $0$ to the maximum number of classes to generate a precision/recall curve, which is common practice in multi-label image classification. 
However, for our proposed DeepSet Network, the number of labels per instance is predicted from the cardinality network. Therefore, prediction/recall is not dependent on value $k$ and one single precision/recall value can be computed.

To calculate the per-class and overall precision/recall, their average values over all classes and all examples are computed, respectively. In addition, we also report the F1 score (the harmonic mean of precision and recall) averaged over all classes (C-F1) and all instances and classes (O-F1).

\begin{figure}[t]
	\centering
	\includegraphics[width=.49\linewidth]{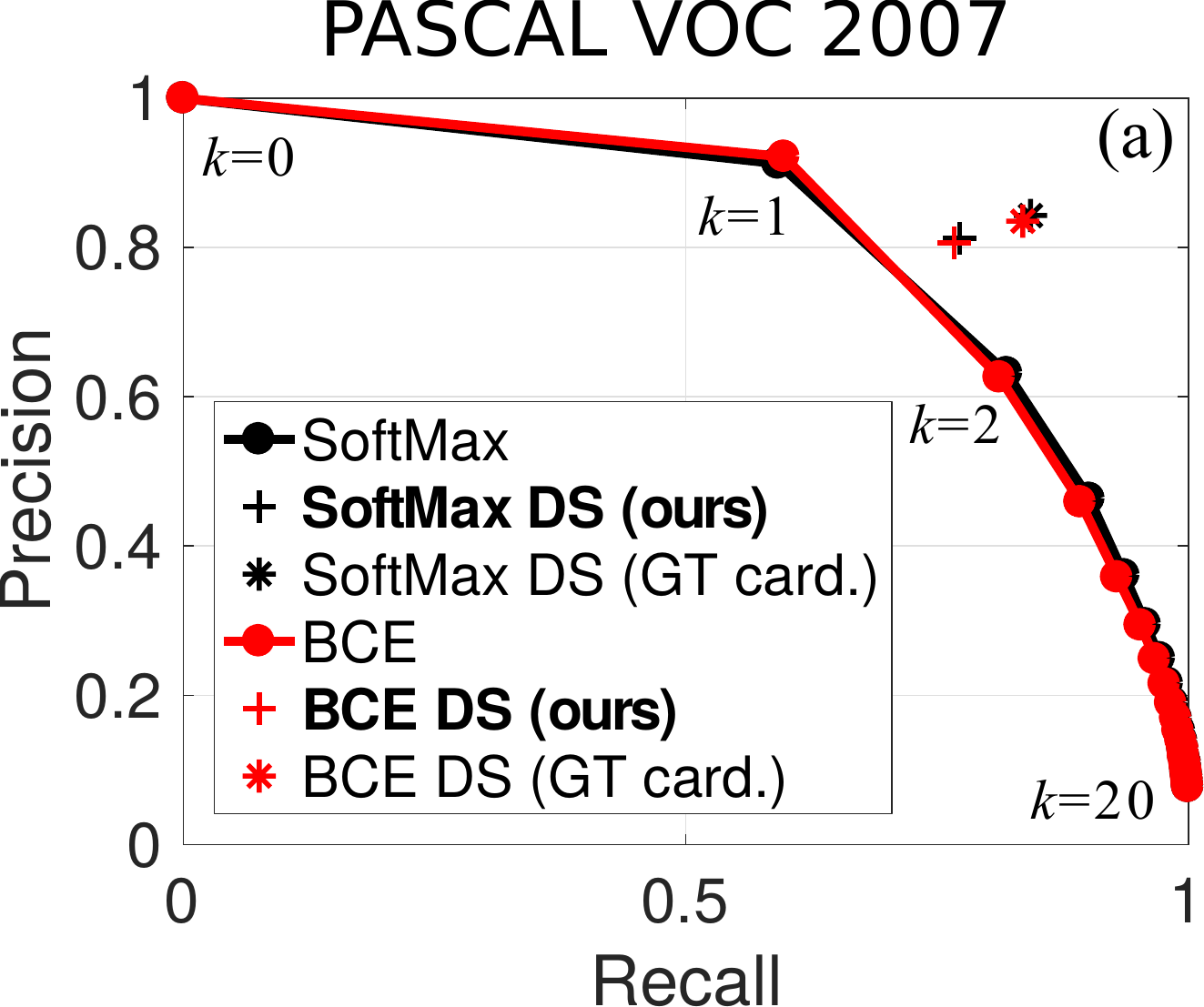}
	\hfill
	\includegraphics[width=.48\linewidth]{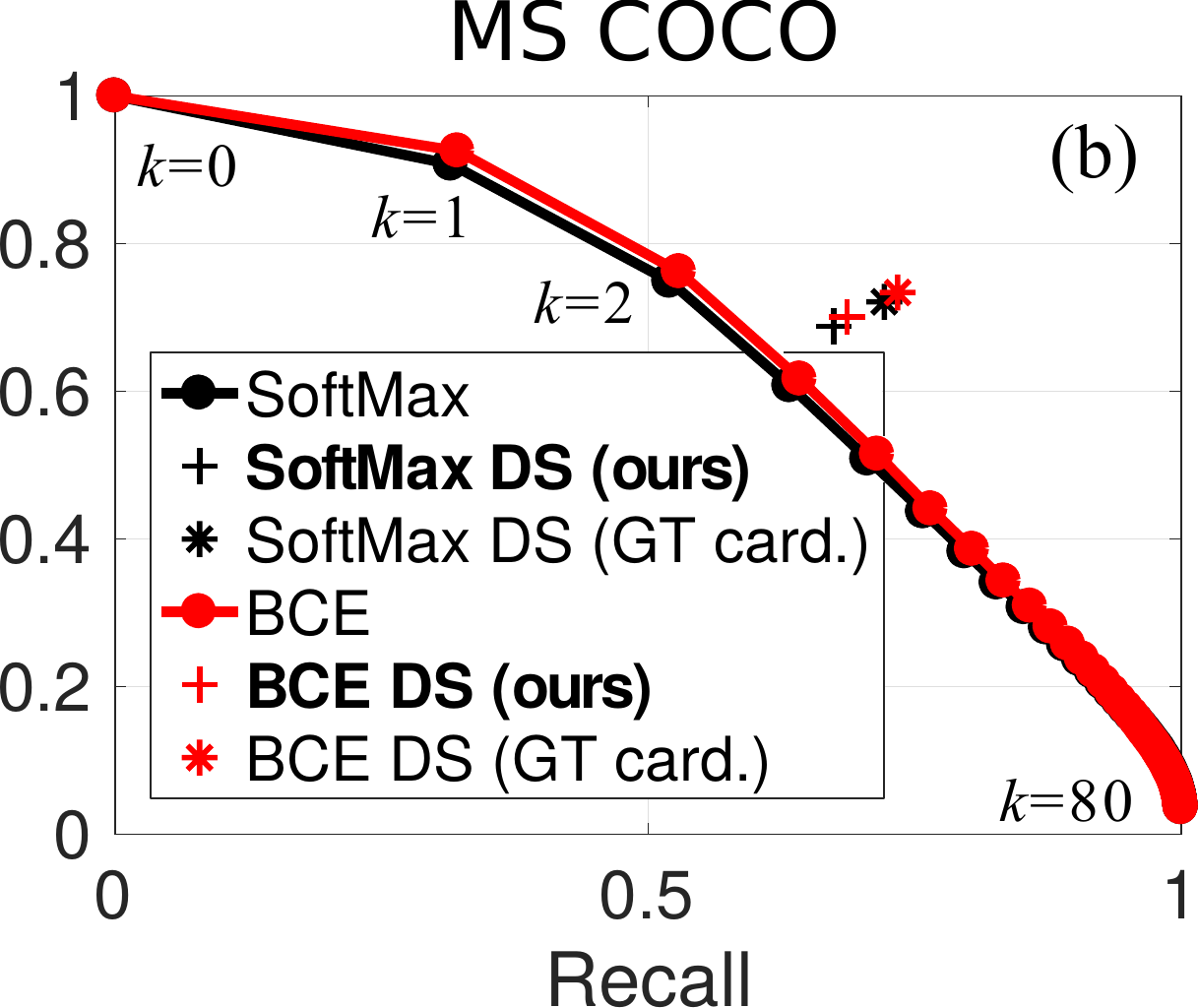}
	\caption{Experimental results on multi-label image classification. The baselines (solid curves) represent state-of-the-art classifiers, fine-tuned for each dataset, using two different loss functions. The methods are evaluated by choosing the top-$k$ predictions across the entire dataset, for different $k$. Our approach predicts $k$ and is thus evaluated only on one single point (+). It outperforms both classifiers significantly in terms of precision and recall and comes very close to the performance when the true cardinality is known ($*$).} 
	\label{fig:curves-mlic}
	\vspace{-1em}
\end{figure}

% \subsubsection{PASCAL VOC 2007}
\myparagraph{PASCAL VOC 2007.} 
The Pascal Visual Object Classes (VOC)~\cite{Everingham:2007:PASCAL-VOC} benchmark is one of the most widely used datasets for detection and classification. It consists of $9963$ images with a 50/50 split for training and test, where objects from $20$ pre-defined categories have been annotated by bounding boxes. Each image may contain between $1$ and $7$ unique objects. 
We compare our results with a state-of-the-art classifier as described above. The resulting precision/recall plots are shown in \Fig~\ref{fig:curves-mlic}(a) together with our proposed approach using the estimated cardinality. Note that by enforcing the correct cardinality for each image, we are able to clearly outperform the baseline \wrt both measures. Note also that our prediction (+) can nearly replicate the oracle ($\ast$), where the ground truth cardinality is known.
The mean absolute cardinality error of our prediction on PASCAL VOC is $0.32 \pm0.52$.

% % \fixmea{Report cardinality estimation error?}\fixmeh{mean absolute error = $0.3241$ and std absolute error  = $0.5235$ on the test set}

%  \begin{figure*}[t]
%  	\centering
%  	\includegraphics[width=.49\linewidth]{figs/PerClass_ROC_Curve_VOC.pdf}
%  	\includegraphics[width=.49\linewidth]{figs/Overal_ROC_Curve_VOC.pdf}
%  	\caption{Pascal VOC 2007.} 
%  	\label{fig:curvesVOC}
%  	\end{figure*}

% \subsubsection{Microsoft COCO} 	
\myparagraph{Microsoft COCO.} 
Another popular benchmark for image captioning, recognition, and segmentation is the  recent Microsoft Common Objects in Context (MS-COCO)~\cite{Lin:2014:COCO}. The dataset consists of
$123$ thousand images, each labelled with per instance
segmentation masks of $80$ classes. The number of unique objects for each image can vary between $0$ and $18$. Around $700$ images in the training dataset do not contain any of the $80$ classes and there are only a handful of images that have more than $10$ tags. The majority of the images contain between one and three labels.  We use $82783$
images as training and validation split ($90$\% - $10\%$), and the remaining $40504$ images as test data. We predict the cardinality of objects in the scene with a mean absolute error of $0.74$ and a standard deviation of $0.86$.

% \begin{table*}[tb]
% %	\footnotesize
% 	\caption{Quantitative results for multi-label image classification on the MS COCO dataset.}
% 	\vspace{-1.5em}
% 	\begin{center}
% 		\begin{tabular}{lc|| ccc| ccc}
% 			\raisebox{-1.0ex}{Classifier}& \raisebox{-1.0ex}{/Metric} & \raisebox{-1.0ex}{C-P} & \raisebox{-1.0ex}{C-R} & \raisebox{-1.0ex}{C-F1} & \raisebox{-1.0ex}{O-P} & \raisebox{-1.0ex}{O-R} & \raisebox{-1.0ex}{O-F1}\\
% 			&\raisebox{-0.5ex}{Evaluation}&&&&&&\\
% 			\hline\hline % inserts single-line
% 			% Entering 1st row
% %			Upper Band&k=3&$64.1$&$74.4$&$68.9$&$75.1$&$78.3$&$76.7$\\ 
% %			\hline
% 			Softmax&k=3&$58.6$&$57.6$&$58.1$ &$60.7$&$63.3$&$62.0$\\
% %			WARP&k=3&&&&&&\\
% 			BCE&k=3&$56.2$&$60.1$&$58.1$&$61.6$&$64.2$&$62.9$\\
% 			CNN-RNN~\cite{Wang_2016_CVPR}&k=3&$66.0$&$55.6$&$60.4$&$69.2$&$66.4$&$67.8$\\
% 			\hline
% 			\textbf{Ours (Softmax)}&Est. Card.&$\textbf{68.2}$&$59.9$&$63.8$&$68.8$&$67.4$&$68.1$\\
% %			\textbf{DeepSetNet (WARP)}&Est. Card.&&&&&&\\
% 			\textbf{Ours (BCE)}&Est. Card.&$66.5$&$\textbf{62.9}$&$\textbf{64.6}$&$\textbf{70.1}$&$\textbf{68.7}$&$\textbf{69.4}$\\
% 
% 		\end{tabular}
% 	\end{center}
% 	\label{table:allcoco-multilabel}
% \end{table*}

%%%%%% ONE COLUMN  ***
\newcommand{\colw}{0.43cm}
\newcommand{\colww}{0.52cm}
\newcommand{\lsh}{\!\!\!\!}
\begin{table}[tb]
	\footnotesize
	\caption{Quantitative results for multi-label image classification on the MS COCO dataset.}
	\vspace{-2em}
	\begin{center}
		\begin{tabular}{lc||p{\colw}p{\colw}p{\colww}| p{\colw}p{\colw}p{\colww} @{}}
% 			\raisebox{-1.0ex}{Classifier}& \raisebox{-1.0ex}{/Metric} & \raisebox{-1.0ex}{C-P} & \raisebox{-1.0ex}{C-R} & \raisebox{-1.0ex}{C-F1} & \raisebox{-1.0ex}{O-P} & \raisebox{-1.0ex}{O-R} & \raisebox{-1.0ex}{O-F1}\\
% 			&\raisebox{-0.5ex}{Evaluation}&&&&&&\\
                        \lsh Classifier & Eval. & \scriptsize{C-P} & \scriptsize{C-R} & \scriptsize{C-F1} & 
                        \scriptsize{O-P} & \scriptsize{O-R} & \scriptsize{O-F1}\\
			\hline\hline % inserts single-line
			% Entering 1st row
%			Upper Band&k=3&$64.1$&$74.4$&$68.9$&$75.1$&$78.3$&$76.7$\\ 
%			\hline
			\lsh Softmax&k=3&$58.6$&$57.6$&$58.1$ &$60.7$&$63.3$&$62.0$\\
%			WARP&k=3&&&&&&\\
			\lsh BCE&k=3&$56.2$&$60.1$&$58.1$&$61.6$&$64.2$&$62.9$\\
			\lsh CNN-RNN~\cite{Wang_2016_CVPR}&k=3&$66.0$&$55.6$&$60.4$&$69.2$&$66.4$&$67.8$\\
			\hline
			\lsh\textbf{Ours (Softmax)}&k=$m^*$ &$\textbf{68.2}$&$59.9$&$63.8$&$68.8$&$67.4$&$68.1$\\
%			\textbf{DeepSetNet (WARP)}&Est. Card.&&&&&&\\
			\lsh\textbf{Ours (BCE)}&k=$m^*$ &$66.5$&$\textbf{62.9}$&$\textbf{64.6}$&$\textbf{70.1}$&$\textbf{68.7}$&$\textbf{69.4}$
		\end{tabular}
	\end{center}
	\label{table:allcoco-multilabel}
						\vspace{-2em}
	\end{table}

\Fig~\ref{fig:curves-mlic}(b) shows a significant improvement of precision and recall and consequently the F1 score using our deep set network compared to the softmax and binary cross-entropy classifiers for all ranking values $k$. We also outperform the state-of-the art multi-label classifier CNN-RNN~\cite{Wang_2016_CVPR}, for the reported value of $k=3$. Our results, listed in \Tab~\ref{table:allcoco-multilabel}, show around $7$ percentage points improvement for the F1 score on top of the baseline classifiers and about $3$ percentage points improvement compared to the state of the art on this dataset. 
Examples of perfect label prediction using our proposed approach are shown in \Fig~\ref{fig:Results1}. The deep set network can properly recognise images with no labels at all, as well as images with many tags. 
We also investigated failure cases where either the cardinality CNN or the classifier fails to make a correct prediction. We showcase some of these cases in Fig~\ref{fig:Results2}. We argue here that some of the failure cases are simply due to a missed ground truth annotation, such as the left-most example, but some are actually semantically correct \wrt the cardinality prediction, but are penalized during evaluation because a particular object category is not available in the dataset. This is best illustrated in the second example in \Fig~\ref{fig:Results2}. Here, our network correctly predicts the number of objects in the scene, which is two, however, the can does not belong to any of the 80 categories in the dataset and is thus not annotated. Similar situations also appear in other images further to the right.

\subsection{Object Counting}
\label{sec:crowd-counting}

% \begin{figure*}[t]
%  	\centering
%  	\includegraphics[width=1\linewidth]{figs/dets}
%  	\caption{Example results of pedestrian detection on selected frames of the MOT16 dataset (cropped for better visibility). Note that we display \emph{all} $m^*$ boxes, and do not threshold on a certain confidence score. The rightmost image shows a failure example, where a wrong cardinality prediction (6 in this case) forces the detector to output many false positives.\fixmea{This figure can go into supp mat if we run out of space}.} 
%  	\label{fig:dets}
%  \end{figure*}
To show the robustness of our cardinality loss, we first evaluate our cardinality estimation on the common crowd counting application. To this end, we apply our approach on the widely used UCSD dataset~\cite{chan2008privacy} and compare our results to four state-of-the art approaches~\cite{arteta2014interactive,onoro2016towards,pham2015count,zhang2015cross}. 
The USCD dataset includes 
a 2000-frames long video sequence, captured by a fixed outdoor surveillance camera. In addition to the video, the region of interest (ROI), the perspective map of the scene and the location annotations of all pedestrians in each frame are also provided.

\myparagraph{Implementation details.} We build our cardinality network structure on top of the well-known AlexNet~\cite{Krizhevsky:2012:NIPS} architecture. However, we replace the first convolutional layer with a single channel filter to accept grayscale images as input, and the last fully connected layer with $2$ layers outputs, similar to the case above (\cf~\Sec~\ref{sec:multi-label}). To estimate the counts, we calculate the mode of the negative binomial distribution. 

As  input, we use a grayscale image constructed by superimposing all region proposals and their scores generated by an off-the-shelf pedestrian detector (before non-maximum suppression). We use the multi-scale deep CNN approach (MS-CNN)~\cite{Cai:2016:ECCV} trained on the KITTI dataset~\cite{Geiger:2012:CVPR} for our purpose. We found, that this input provides a stronger signal than the raw RGB images, yielding better results. 
Note that we process the input images with a pedestrian detector, however, we do not use any location or perspective information that is available for this dataset. During learning, we only rely on the object count for each image region.

% \fixmea{TODO} Although we use different input compared to most crowd counting approaches, we do not predict a density map and therefore, we do not use the dataset's location annotations and/or perspective map for training our network. During learning, we only provide the ground truth counts for our cardinality estimation network and use the loss introduced in Eq.~\eqref{eq:Cardinal_Eq0}.  

\renewcommand{\colw}{0.43cm}
\renewcommand{\colww}{0.64cm}
\renewcommand{\lsh}{\!\!\!\!}
\begin{table}[tb]
	\footnotesize
	\caption{Count mean absolute error on UCSD dataset.}
	\vspace{-2em}
		\begin{center}
		\begin{tabular}{l|ccccc}
			\lsh Method& max & downscale & upscale & min & overall\\
			\hline\hline
			\lsh C-Forest~\cite{pham2015count}&$1.43$&$1.30$&$1.59$&$1.62$&$1.49$\\
			\lsh IOC~\cite{arteta2014interactive}&$1.24$&$ 1.31$&$ 1.69$&$\textbf{1.49}$&$\textbf{1.43}$\\
			\lsh Cs-CCNN~\cite{zhang2015cross}& $1.70$ & $\textbf{ 1.26}$ & $1.59$ & $1.52$ & $ 1.52$ \\
			\lsh CCNN~\cite{onoro2016towards} & $1.65$ & $1.79$ & $1.11$ & $1.50$ & $1.51$\\
			\hline
			\lsh Hydra 2s~\cite{onoro2016towards} & $2.22$ & $1.93$ & $ 1.37$ & $2.38$ & $1.98$\\
			\lsh Hydra 3s~\cite{onoro2016towards} & $2.17$ & $ 2.99$ & $1.44$ & $1.92$ & $2.13$\\
			\hline 
			\lsh \textbf{Ours} & $\textbf{1.23}$ & $1.60$  & $\textbf{0.79}$ & $2.62$ & $1.56$ \\

		\end{tabular}
	\end{center}
	\label{table:Crowd_err}
		\vspace{-2.5em}
\end{table}
We follow exactly the same data split used in~\cite{onoro2016towards} by conducting four different and separate experiments on maximal, downscale, upscale and minimal subsets in UCSD dataset. 
 In order to train our network, similar to~\cite{onoro2016towards} we use data augmentation in each experiment by extracting $800$ random patches from each training image and their corresponding
ground truth counts. We also randomly flip each patch during training. To ensure that we can count all pedestrians from the entire image at test time, we choose the patch sizes to be exactly half of the image size ($79\times119$ pixels) and then perform inference on the resulting $4$ non-overlapping regions. The weights are initialised randomly and the network is trained for $100$ epochs.
All hyperparameters are set as in \Sec~\ref{sec:multi-label}.
% We set a constant learning rate $10^{-3}$, weight decay $5\cdot 10^{-4}$, momentum rate $0.9$ and dropout rate $0.5$. The checkpoint with the lowest objective on the validation set is chosen for prediction.

\myparagraph{Results.} \Tab~\ref{table:Crowd_err} shows the mean absolute error between the predicted and the ground truth counts. We show competitive or superior performance in each experiment except for the `minimal' subset. The main reason is that  the training set size is too small (only $10$ images) in this particular split and even data augmentation cannot generalize the cardinality model for the test images. Moreover, unlike other methods, we do not utilize any location information but only provide the object count as ground truth.
% we provide very compact information for our network about the counts while other approaches use density map which could be helpful to learn a better model for such small subset. 
Considering the overall performance, our approach outperforms state-of-the-art counting approaches  that do not use the perspective map (Hydra 2s and 3s) and performs favourably compared to many existing methods that exploit localisation and perspective information.
% . We also perform comparably well in overall performance compared to the rest of crowd counting approaches which use additional information such as the perspective map and ground truth annotation to learn their density map.  

% \fixmel{This doesn't really match the state of art claim in abstract, intro...}

\myparagraph{Discussion.}
One obvious alternative for our proposed cardinality loss may seem to directly regress for $m$. This alternative, however, has two main drawbacks. First, it cannot be formulated within a Bayesian set framework to model uncertainty, and second, the regression loss does not yield a discrete distribution and hence does not fit the underlying mathematical foundation of this paper. Nevertheless, we have run the same experiments as above using a standard regression loss but did not reach the same performance.
Using the regression loss we achieve a mean cardinality error (MCE) of $0.83$ on MS COCO, while our loss yields an MCE of $0.74$. This is also reflected in the O-F1 score which drops from $69.4$ to $68.4$ when directly regressing for $m$.

% Another practical alternative to our cardinality loss is the commonly used regression loss. Although regressing for cardinality may seem obvious, it does not give best results in our experiments. Moreover, this approach is inferior to our derived loss because it
% a) cannot be accommodated in a Bayesian set formulation to model uncertainty and 
% b) does not yield a discrete distribution. Therefore, it does not fit the mathematics in this paper. Our cardinality loss in addition to its acceptable performance provides a distribution for the cardinality, which can be used for Bayesian analysis of the problem.

%%%%%%%%%%%%%%%%%%%%%%%%%%%%%%%%%%%%%%%%%%%%%%%%%%%%%%%%%%%%
\subsection{Pedestrian Detection}
\label{sec:detection}
\begin{figure}[t]
\centering
\begin{subfigure}[t]{.31\linewidth}
    \centering
    \includegraphics[width=1\linewidth]{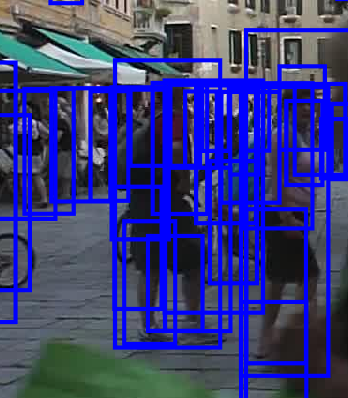}
    \caption{Proposals}
\end{subfigure}
\hfill
\begin{subfigure}[t]{.31\linewidth}
    \centering
    \includegraphics[width=1\linewidth]{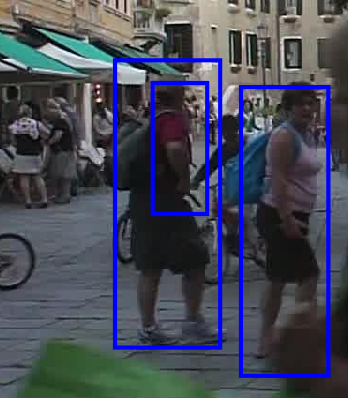}
    \caption{MS-CNN~\cite{Cai:2016:ECCV}}
\end{subfigure}
\hfill
\begin{subfigure}[t]{.31\linewidth}
    \centering
    \includegraphics[width=1\linewidth]{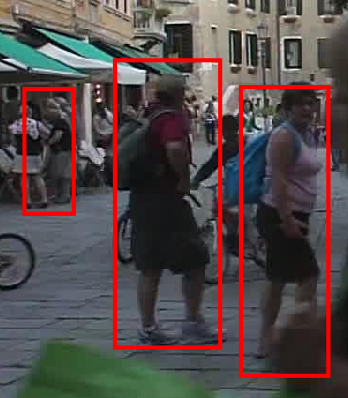}
    \caption{Our result}
\end{subfigure}
\vspace{-.5em}
% 'MOT16-RegProps', 'MOT16-ours', 'MOT16-MS-CNN'
\small
\caption{Example pedestrian detection result of our approach. To select relevant detection candidates from an overcomplete set of proposals~\ep{a}, state-of-the-art methods rely on non-maximum suppression (NMS) with a fixed setting~\ep{b}. We show that a better result can be achieved by adjusting the NMS threshold adaptively, depending on the number of instances in each image (3 in this case)~\ep{c}.}
\vspace{-1.5em}
\label{fig:nms}
\end{figure}
% In this application, we aim to demonstrate that incorporating cardinality prediction (person count) can be beneficial to the task of pedestrian detection, which is naturally a set prediction problem. 
In this section, we cast the task of pedestrian detection as a set prediction problem and demonstrate that incorporating cardinality prediction (person count) can be beneficial to improve performance.
% Combination of the cardinality estimations and the detectors will also support our set formulation in \Eq~\eqref{eq:set_density}, which consists of two terms of cardinality (counting) and spatial (pedestrian detector) distribution.   %% I'm not sure if it belongs here.
To this end, we perform experiments on two widely used datasets, Caltech Pedestrians~\cite{Dollar:2012:PAMI} and MOT16 from the MOTChallenge benchmark~\cite{Milan:2016:MOT16}. Recalling \Eqs~\eqref{eq:CNN_Eq} and~\eqref{eq:Cardinal_Eq0}, we need two networks with parameters $\bw^*$ and $\btheta^*$ for cardinality estimation and detection proposals, respectively. For the cardinality network, we use the exact same architecture and setup as in \Sec~\ref{sec:crowd-counting} and train it on the training sets of these datasets.
% \footnote{More details on implementation and dataset split and further results are provided in supplementary material.} % removed due to space limit
%\textcolor{red}{
%Although we could learn our own pedestrian detector $\btheta^*$, to push the state-of-the art performance on pedestrian detection, we chose the leading detector on Caltech Pedestrians, which is the recent the multi-scale deep CNN approach (MS-CNN)~\cite{Cai:2016:ECCV}.}
Note that it is not our intention to engineer a completely novel pedestrian detector here. Rather, for $\btheta^*$, we take an off-the-shelf state-of-the-art system (MS-CNN)~\cite{Cai:2016:ECCV} and show how it can be further improved by taking the cardinality prediction into account.

To generate the final detection outputs, most detectors often rely on non-maximum suppression (NMS), which greedily picks the boxes with highest scores and suppresses any boxes that overlap more than a pre-defined threshold $T_O$. This heuristic makes the solution more ad-hoc than what is expressed in our set formulation in \Eq~\eqref{eq:set_density}. However, we are still able to improve the detector performance by adjusting this threshold for each frame separately. To obtain the final detection output, we use the prediction on the number of people $(m^*)$ in the scene to choose an adaptive NMS threshold for each image. In particular, we start from the default value of $T_O$, and increase it gradually until the number of boxes reaches $m^*$. In the case if the number of final boxes is larger than $m^*$, we pick $m^*$ boxes with the highest scores, which corresponds to the MAP set prediction as discussed in \Sec~\ref{sec:inference}. To ensure a fair comparison, we also determine the best (global) value for $T_O=0.4$ for the MS-CNN baseline. \Fig~\ref{fig:nms} demonstrates an example of the adjusted NMS threshold when considering the number of pedestrians in the image.

To quantify the detection performance, we adapt the same evaluation metrics and follow the protocols used on the Caltech detection benchmark~\cite{Dollar:2012:PAMI}. The evaluation metrics used here are log-average
miss rate (MR) over false positive per image. Additionally, we compute the F1 score (the harmonic mean of precision and recall). The F1 score is computed from \emph{all} detections predicted from our DeepSet network and is compared with the \emph{highest} F1 score along the MS-CNN precision-recall curve. To calculate MR, we concatenate all boxes resulted from our adaptive NMS approach and change the threshold over all scores from our predicted sets. 
Quantitative detection results are shown in \Tab~\ref{table:F1 score}. Note that we do not retrain the detector, but are still able to improve its performance by predicting the number of pedestrians in each frame in these two dataset.

\newcommand{\shft}{\hspace{-.2cm}}
\begin{table}[tb]
	\small
	\caption{Pedestrian detection results measured by F1 score (higher is better) and log-average miss rate (lower is better).}
\vspace{-2em}	
	\begin{center}
		\begin{tabular}{lcccc}
\toprule
			& \multicolumn{2}{c}{F1-score  $\uparrow$ } & \multicolumn{2}{c}{MR $\downarrow$} \\
			\cmidrule(l{2pt}r{2pt}){2-3}  \cmidrule(l{2pt}r{2pt}){4-5} 
\shft 			Method & Caltech & \shft MOT16 & Calt. & \shft MOT16\\
			\midrule
\shft 			MS-CNN~\cite{Cai:2016:ECCV} & $51.61$ & $59.04$ & $60.9$ & $82.8$\\
			% \hline
\shft 			\bf MS-CNN-DS (ours)& $52.15$ & $61.86$ & $60.4$ & $81.7$ \\
\shft 		    MS-CNN-DS (GT card.) & $52.28$ & $62.42$ & $60.3$ & $81.5$ \\
\bottomrule
		\end{tabular}
	\end{center}
	\label{table:F1 score}
	\vspace{-2.5em}	
\end{table}

%%%%%%%%% Figure Ped Detection curves %%%%%%%%%%%
% \begin{figure}[ht]
% \centering
% \includegraphics[width=.9\linewidth]{figs/UsaTestRocAll.pdf}
% \caption{Pedestrian detection results on the Caltech benchmark test set.}
% \label{fig:detection-caltech}
% \end{figure}

% %%%%%%%%% Figure global M %%%%%%%%%%
% \begin{figure}[ht]
% \centering
% \includegraphics[width=.9\linewidth]{figs/globalM_MOT16_test.pdf}
% \caption{Choosing a global threshold.}
% \label{fig:global_trehshold}
% \end{figure}

\section{Conclusion}
\label{sec:conclusion}

We proposed a deep learning approach for predicting sets. To achieve this goal, we derived a loss for learning a discrete distribution over the set cardinality.
This allowed us to use standard backpropagation for training a deep network for set prediction.
We have demonstrated the effectiveness of this approach on crowd counting, pedestrian detection and multi-class image classification, achieving competitive or superior results in all three applications. As our network is trained independently, it can be trivially applied to any existing classifier or detector, to further improve performance. 

Note that this decoupling is a direct consequence of our underlying mathematical derivation due to the \iid~assumptions, which renders our approach very general and applicable to a wide range of models.
In future, we plan to extend our method to multi-class cardinality estimation and investigate models that do not make \iid assumptions. Another potential avenue could be to exploit the Bayesian nature of the model to include uncertainty as opposed to relying on the MAP estimation alone. 

\myparagraph{Acknowledgments.}
This research was supported by the Australian Research Council through the Centre of Excellence in Robotic Vision, CE140100016, and through Laureate Fellowship FL130100102 to IDR.
%------------------------------------------------------------------------

{\small
	\bibliographystyle{ieee}
	\bibliography{reference/ref,reference/refs-short,reference/anton-ref}
}
\onecolumn
\appendix

%%%%%%%%% BODY TEXT
\section*{Appendix}

%%%%%%%%% BODY TEXT

This appendix accompanies the main text. We first provide more background on finite set statistics. Further, we add the details of our derivations for Deep Set Network that were omitted due to space constraints. To do this, here we augment Sections $3$ and $4$ of the main text.  
Finally, we provide more discussions and results on all object counting, multi-label classification and pedestrian detection applications. 
%-------------------------------------------------------------------------
\section{Background on Finite Set Statistics}

\label{sec:background0}
Finite Set Statistics  provide powerful and practical mathematical tools for dealing with
random finite sets, based on the notion of integration and density that is consistent with the point process theory~\cite{mahler2007statistical}. In this section, we review some basic mathematical background about this subject of statistics. 

In the conventional statistics theory, a continuous random variable $y$ is a variable that can take an infinite number of possible values. A continuous random vector can be defined by stacking several continuous random variables into a fixed length vector, $Y=\left(y_1,\cdots,y_m\right)$. The mathematical function describing the possible values of a continuous random vector, and their associated joint probabilities, is known as a probability density function (PDF) $p(Y)$ such that
$\int p(Y)dY = 1.$
%**********************************************************************************************
% \begin{equation}\label{eq: pdf}
% \begin{aligned}
% %\int_{-\infty}^{\infty} p(Y)dY = 1.
% \int p(Y)dY = 1.
% \end{aligned}
% \end{equation}
%**********************************************************************************************

A random finite set (RFS) $\calY$ is a finite-set valued random variable $\calY=\left\{y_1,\cdots,y_m\right\}\subset \mathbb{Y}$. The main difference between an RFS and a random vector is that for the former, the number of constituent variables, $m$, is random and the variables themselves are random and unordered, while the latter is of a fixed size with a predefined order.

A statistical function describing a finite-set variable $\calY$ is a
combinatorial probability density function $p(\calY)$, which consists of a discrete probability distribution, the so-called cardinality distribution, and a family of joint probability densities on  the values of the constituent variables for each cardinality. 
Similar to the definition of a PDF for a random variable, the PDF of an RFS must sum to unity over all possible cardinality values and all possible element values and their permutations, \ie
%**********************************************************************************************
\begin{equation}\label{eq: RFS pdf0}
\int p(\calY)\mu(d\calY) \triangleq \sum_{m=0}^{\infty}\frac{1}{m!U^m}\int p(\{y_1,\cdots,y_m\}_{||}) dy_1\cdots dy_m = 1,
\end{equation}
%**********************************************************************************************
where $\mu$ is the dominating
measure and $U$ is the unit of hypervolume
in $\mathbb{Y}$~\cite{vo2016model}. 
The PDF of an $m$-dimensional random vector can be defined in terms of an RFS as:
%**********************************************************************************************
\begin{equation}\label{eq: pdf rfs vs vect0}
\begin{aligned}
p(y_1,\cdots,y_m) \triangleq \frac{1}{m!} p(\{y_1,\cdots,y_m\}_{||}).
\end{aligned}
\end{equation}
%**********************************************************************************************
The denominator $m!=\prod_{k=1}^m k$ appears because the probability density for a set $\{y_1,\cdots,y_m\}_{||}$ must be equally distributed among all the $m!$ possible permutations of the vector~\cite{mahler2007statistical}.

The cardinality distribution $p(m)$ over the number of
elements in the random finite set $\calY$ is obtained by
%**********************************************************************************************
\begin{equation}\label{eq: Cardinality distribution}
p(m) = \int_{|\calY|=m} p(\calY)\mu(d\calY) \triangleq \frac{1}{m!U^m} \int p(\{y_1,\cdots,y_m\}_{||}) dy_1\cdots dy_m.
\end{equation}
%**********************************************************************************************
Similar to the conventional statistics for random variables, the expectation of an RFS has been defined above.  
The first statistical moment, or the expected value, of an RFS is known as intensity density or probability hypothesis density (PHD) and is calculated by definition as 
%**********************************************************************************************
\begin{equation}\label{eq: intensity density}
v(y) \triangleq  \int\delta_{\calY}(y) p(\calY)\mu(d\calY),
\end{equation}
%**********************************************************************************************
where $\delta_{\calY}(y) = \sum_{x\in \calY}\delta_x(y)$ and $\delta_x(y)$ denotes the Dirac delta function
concentrated at $x$. The PHD function $v(y)$ is interpreted as the instantaneous expected number of the variables that exist
at that point $y$. Moreover, the integral of the PHD over a region gives the expected number of elements in that region and the peaks of the PHD
indicate highest local concentrations of the expected number of elements.

Having an RFS distribtuion $p(\calY)$, the samples can be drawn from this distribution as shown in Algorithm~\ref{table:RFS}. 
%The mode of this distribution is defined as the mode of the cardinality distribution $p(m)$ and the mode of the joint distribution for $m^*$ elements, \textit{i.e.} $p(\{y_1,\cdots,y_{m^*}\})$ .    

\renewcommand{\tablename}{Algorithm}
\begin{table}[tb]
\caption{Drawing samples from a set distribution.}
\centering
		\begin{tabular}{l}
			\hline
             \textbf{Sampling an RFS Probability Distribution}\\
             \hline\hline
			\tabitem Initialize $\calY \leftarrow \emptyset$\\
			\tabitem Sample cardinality $m \sim p(m)$\\
			\tabitem Sample $m$ points from an $m$-dimensional joint distribution\\
			\qquad$\calY \sim p(\{y_1,y_2,\cdots.y_m\}_{||}) \leftarrow m!\times p(y_1,y_2,\cdots.y_m)$\\
			\quad \textit{In the case of i.i.d. samples:}\\
			\qquad for $i \leftarrow \{1,\ldots,m\}$\\
			\qquad \qquad sample $y_i \sim p(y) $\\
			\qquad \qquad set $\calY \leftarrow \calY\cup y_i $\\
			\qquad end\\ 
			\hline
		\end{tabular}	
	\label{table:RFS}
\end{table}
\renewcommand{\tablename}{Table}

\section{Deep Set Network}
\label{sec:deep-set-net0}

Let us begin by defining a training set $\D = \{\calY_{i},\bx_{i}\}$,
% \begin{eqnarray*}
% 	\calY & = & \{\by_{1},\by_{2},\ldots,\by_{m}\}\qquad \by_{k}\in\mathbb{R}^{d}, \forall k\\ % \text{or}\quad\mathbb{Z}^{d} \\
% 	\D & = & \{\calY_{i},\bx_{i}\}\qquad\qquad\quad\quad\forall\bx_{i}\in\mathbb{R}^{l}\quad\forall i=1,\ldots,n,
% \end{eqnarray*}
where each training sample $i=1,\ldots,n$ is a pair consisting of an input feature $\bx_{i}\in\mathbb{R}^{l}$ and an output (or label) set
$\calY_{i} = \{y_{1},y_{2},\ldots,y_{m_i}\}, y_{k}\in\mathbb{R}^{d}, m_i\in\mathbb{N}^0 $. In the following we will drop the instance index $i$ for better readability. Note that $m:=|\calY|$ denotes the cardinality of set $\calY$.
The probability of a set $\calY$ with an unknown cardinality is defined as:
\begin{equation}
\label{eq:set_density0}
\begin{aligned}
p(\calY|\bx,\btheta,\bw) = & p(m|\bx,\bw)\times U^m\times p(\{y_{1},y_{2},\cdots,y_{m}\}_{||}|\bx,\btheta)\\
= & p(m|\bx,\bw)\times m!\times U^m\times p(y_{1},y_{2},\cdots,y_{m}|\bx,\btheta),
\end{aligned}
\end{equation} 
where $p(m|\cdot,\cdot)$ and $ p(y_{1},y_{2},\cdots,y_{m}|\cdot,\cdot)$ are respectively a cardinality distribution and a
symmetric joint probability density distribution of the elements. $U$ is the unit of hypervolume
in $\mathbb{R}^{d}$, which makes the joint distribution unitless~\cite{vo2016model}. $\btheta$ denotes the parameters that estimate the joint distribution of set element values for a fixed cardinality,
while $\bw$ represents the collection of parameters which estimate the cardinality distribution of the set elements. 

The above formulation represents the probability density of a set which is very general and completely independent from the choices of both cardinality and spatial distributions. It is thus straightforward to transfer it to many applications that require the output to be a set.  However, to make the problem amenable to mathematical derivation and implementation, we adopt two assumptions: \emph{i)} the outputs (or labels) in the set are independent
and identically distributed (\iid) and  \emph{ii)} their cardinality follows
a Poisson distribution with parameter $\lambda$. Thus, we can write the distribution as
\begin{equation}
\begin{aligned}
p(\calY|\bx,\btheta,\bw) = \int p(m|\lambda)p(\lambda|\bx,\bw) d\lambda\times
m!\times U^m\times\left(\prod_{k=1}^{m}p(y_{k}|\bx,\btheta)\right).
\end{aligned}   
\end{equation}

\subsection{Posterior distribution}
\label{sec:posterior}
To learn the parameters $\btheta$ and $\bw$, it is valid to assume that the training samples are independent from each other and the distribution over the input data $p(\bx)$ is independent from the output and the parameters. 
Therefore, the posterior distribution over the parameters can be derived as
\begin{equation}
\begin{aligned}
p(\btheta,\bw|\D) &= \frac{1}{Z} p(\D|\btheta,\bw)p(\btheta)p(\bw)\\
&= \frac{1}{Z} p(\{\calY_{i},\bx_{i}\}_{\forall i}|\btheta,\bw)p(\btheta)p(\bw)\\
&= \frac{1}{Z}\prod_{i=1}^{n} \bigg[p(\calY_{i}|\bx_{i},\btheta,\bw) p(\bx_{i})\bigg]p(\btheta)p(\bw)\\
&= \frac{1}{Z}\prod_{i=1}^{n}\left[\int p(m_{i}|\lambda)p(\lambda|\bx_{i},\bw)d\lambda\times 
m_{i}!\times U^{m_i}\times\left(\prod_{k=1}^{m_{i}}p(y_{k}|\bx_{i},\btheta)\right) p(\bx_{i})\right]p(\btheta)p(\bw),
\end{aligned} 
\label{eq:posterior}
\end{equation}
where $Z$ is a normaliser defined as 
\begin{equation}
Z = \int \int\prod_{i=1}^{n} \left[\int p(m_{i}|\lambda)p(\lambda|\bx_{i},\bw)d\lambda\times 
m_{i}!\times U^{m_i}\times\left(\prod_{k=1}^{m_{i}}p(y_{k}|\bx_{i},\btheta)\right) p(\bx_{i})\right]p(\btheta)p(\bw)\quad d\theta d\bw.
\end{equation}
The probability
$p(\bx_{i})$ can be eliminated as it appears in both the numerator and the denominator. Therefore,  
\begin{equation}
p(\btheta,\bw|\D) = \frac{1}{\tilde{Z}}\prod_{i=1}^{n}\left[\int p(m_{i}|\lambda)p(\lambda|\bx_{i},\bw)d\lambda\times 
m_{i}!\times U^{m_i}\times\left(\prod_{k=1}^{m_{i}}p(y_{k}|\bx_{i},\btheta)\right)\right]p(\btheta)p(\bw),
\label{eq:posterior_m}
\end{equation}
where 
\begin{equation}
\tilde{Z} = \int \int\prod_{i=1}^{n} \left[\int p(m_{i}|\lambda)p(\lambda|\bx_{i},\bw)d\lambda\times 
m_{i}!\times U^{m_i}\times\left(\prod_{k=1}^{m_{i}}p(y_{k}|\bx_{i},\btheta)\right)\right]p(\btheta)p(\bw)\quad d\theta d\bw.
\end{equation}

A closed form solution for the integral in \Eq~\eqref{eq:posterior_m} can be obtained by using conjugate priors:
\begin{eqnarray*}
	m & \sim & \mathcal{P}(m;\lambda)\\
	\lambda & \sim & \mathcal{G}(\lambda;\alpha(\bx,\bw),\beta(\bx,\bw))\\
	&&\alpha(\bx,\bw),\beta(\bx,\bw)>0\quad\forall\bx,\bw\\
	\btheta & \sim & \mathcal{N}(\btheta;0,\sigma_{1}^{2}\mathbf{I})\\
	\bw & \sim & \mathcal{N}(\bw;0,\sigma_{2}^{2}\mathbf{I}),
\end{eqnarray*}
where $\mathcal{P}(\cdot,\lambda)$, $\mathcal{G}(\cdot;\alpha,\beta)$,  and $\mathcal{N}(\cdot;0,\sigma^{2}\mathbf{I})$ represent respectively a Poisson distribution with parameters $\lambda$, a Gamma distribution with parameters $(\alpha,\beta)$ and a zero mean normal distribution with covariance equal to $\sigma^{2}\mathbf{I}$. 

We assume that the cardinality follows a Poisson distribution whose mean, $\lambda$, follows a Gamma distribution, with parameters which can be estimated from the input data $\bx$. 
Note that the cardinality distribution in \Eq~\ref{eq:set_density0} can be replaced by any other discrete distribution. For example, it is a valid assumption to model the number of objects in natural images by a Poisson distribution~\cite{chan2009bayesian}. Thus, we could directly predict $\lambda$ to model this distribution by formulating the cardinality as $p(m|\bx,\bw) = \mathcal{P}(m;\lambda(\bx,\bw))$ . However, this would limit the model's expressive power; because two visually entirely different images with the same number of objects would be mapped to the same $\lambda$. Instead, to allow for uncertainty of the mean, we model it with another distribution, which we choose to be Gamma for mathematical convenience.  
Consequently, the integrals in $p(\btheta,\bw|\D)$ are simplified
and form a negative binomial distribution,
\begin{equation}
\NB\left(m;a,b\right)  = \frac{\Gamma(m+a)}{\Gamma(m+1)\Gamma(a)}.(1-b)^{a}b^{m}, 
\end{equation}
where $\Gamma$ is the Gamma function. Finally, the full posterior distribution can be written as

\begin{equation}
\begin{aligned}
p(\btheta,\bw|\D)  =\frac{1}{\tilde{Z}}\prod_{i=1}^{n}\bigg[\NB\left(m_{i};\alpha(\bx_{i},\bw),\frac{1}{1+\beta(\bx_{i},\bw)}\right)\times m_{i}!\times U^{m_{i}}\times\left(\prod_{k=1}^{m_{i}}p(y_{k}|\bx_{i},\btheta)\right)\bigg]p(\btheta)p(\bw).
\label{eq:full-posterior}
\end{aligned}   
\end{equation}

\subsection{Learning}
\label{sec:learning}

For simplicity, we use a point estimate for the posterior $p(\btheta,\bw|\D)$,
\ie $p(\btheta,\bw|\D) = \delta(\btheta=\btheta^{*},\bw=\bw^{*}|\D)$, where $(\btheta^{*},\bw^{*})$ are computed using the following MAP estimator: 
\begin{equation}
(\btheta^{*},\bw^{*}) = \arg\max_{\btheta,\bw}\quad \log\left(p\left(\btheta,\bw|\D\right)\right).
\label{eq:map}
\end{equation}

Since the solution to the above problem is independent from the normalisation constant $\tilde{Z}$, we have  
\begin{equation}
\begin{aligned}
(\btheta^{*},\bw^{*})=  \arg\max_{\btheta,\bw}&\quad\log\left(p(\btheta)\right)+\sum_{i=1}^{n}\left[\log\left(m_{i}!\right)+m_i\log U+\sum_{k=1}^{m_{i}}\log\left(p(y_{k}|\bx_{i},\btheta)\right)\right.\\
& \quad+\left.\log\left(NB\left(m_{i};\alpha(\bx_{i},\bw),\frac{1}{1+\beta(\bx_{i},\bw)}\right)\right)\right]+\log\left(p(\bw)\right)\\
= & \arg\max_{\btheta,\bw}\quad f_{1}(\btheta)+f_{2}(\bw).
\end{aligned}
\label{eq:map_complete}
\end{equation}

Therefore, 
the optimisation problem in Eq.~\eqref{eq:map_complete} can be decomposed \wrt the parameters
$\btheta$ and $\bw$. Therefore, we can learn them independently in two separate problems

\begin{equation}
\begin{aligned}
\btheta^{*} = & \arg\max_{\btheta}\quad f_{1}(\btheta)\\
= & \arg\max_{\btheta}\quad\gamma_{1}\|\btheta\|+\sum_{i=1}^{n}\left[\log\left(m_{i}!\right)+m_i\log U+\sum_{k=1}^{m_{i}}\log\left(p(y_{k}|\bx_{i},\btheta)\right)\right]\\
\equiv&\arg\max_{\btheta}\quad\gamma_{1}\|\btheta\|+\sum_{i=1}^{n}\sum_{k=1}^{m_{i}}\log\left(p(y_{k}|\bx_{i},\btheta)\right)
\end{aligned}
\label{eq:CNN_Eq0}
\end{equation}
%or equivalently 
%\begin{equation}
%\btheta^{*} =  \arg\max_{\btheta}\quad\gamma_{1}\|\btheta\|+\sum_{i=1}^{n}\sum_{k=1}^{m_{i}}\log\left(p(\by_{k}|\btheta,\bx_{i})\right)
%\label{eq:CNN_Eq}
%\end{equation}
and 
\begin{equation}
\begin{aligned}
\bw^{*} =  \arg\max_{\bw}\quad& f_{2}(\bw)\\
=  \arg\max_{\bw}\quad&\sum_{i=1}^{n}\Big[\log\left(\frac{\Gamma(m_{i}+\alpha(\bx_{i},\bw))}{\Gamma(m_{i}+1)\Gamma(\alpha(\bx_{i},\bw))}\right)\\
& +  \displaystyle{ \log\left(\frac{\beta(\bx_{i},\bw)^{\alpha(\bx_{i},\bw)}}{(1+\beta(\bx_{i},\bw)^{\alpha(\bx_{i},\bw)+m_{i}})}\right)}\Big]+\gamma_2\|\bw\|,
\label{eq:Cardinal_Eq}
\end{aligned}   
\end{equation}
where $\gamma_1$ and $\gamma_2$ are the regularisation parameters, proportional to the predefined covariance parameters $\sigma_1$ and $\sigma_2$. These parameters are also known as weight decay parameters and commonly used in training neural networks.

The learned parameters $\btheta^{*}$ in \Eq~\eqref{eq:CNN_Eq0} are used to map an input feature vector $\bx$  into an output vector $Y$. For example, in image classification, $\btheta^*$ is used to predict the distribution $Y$ over all categories, given the input image $\bx$. Note that $\btheta^*$ can generally be learned using a number of existing machine learning techniques. In this paper we rely on deep CNNs to perform this task.

%To learn the highly complex function between the input feature $\bx$ and the parameters $(\alpha,\beta)$, which are used for estimating the output cardinality distribution, we train a second deep neural network.
%
To learn the highly complex function between the input feature $\bx$ and the parameters $(\alpha,\beta)$, which are used for estimating the output cardinality distribution, we train a second deep neural network. 
Using neural networks to predict a discrete value may seem counterintuitive, because these methods at their core rely on the backpropagation algorithm, which assumes a differentiable loss. Note that we achieve this by describing the discrete distribution by continuous parameters $\alpha, \beta$ (Negative binomial $\NB(\cdot,\alpha,\frac{1}{1+\beta})$), and can then easily draw discrete samples from that distribution. More formally, to estimate $\bw^{*}$, we compute the partial derivatives of the objective function in \Eq~\eqref{eq:Cardinal_Eq} \wrt $\alpha (\cdot,\cdot)$ and $\beta (\cdot,\cdot)$ and use standard backpropagation to learn the parameters of the deep neural network. 
\begin{equation}
\frac{\partial f_{2}(\bw)}{\partial\bw} = \frac{\partial f_{2}(\bw)}{\partial\alpha(\bx,\bw)}.\frac{\partial\alpha(\bx,\bw)}{\partial\bw}+\frac{\partial f_{2}(\bw)}{\partial\beta(\bx,\bw)}.\frac{\partial\beta(\bx,\bw)}{\partial\bw}+2\gamma_{2}\bw,
\end{equation}
where 
\begin{equation}
\frac{\partial f_{2}(\bw)}{\partial\alpha(\bx,\bw)} =  \sum_{i=1}^{n}\bigg[\Psi\Big(m_{i}+\alpha(\bx_{i},\bw)\Big)-\Psi\Big(\alpha(\bx_{i},\bw)\Big)+\log\Big(\frac{\beta(\bx_{i},\bw)}{1+\beta(\bx_{i},\bw)}\Big)\bigg],
\end{equation}
and 
\begin{equation}
\frac{\partial f_{2}(\bw)}{\partial\beta(\bx,\bw)} =  \sum_{i=1}^{n}\bigg[\frac{\alpha(\bx_{i},\bw)-m_{i}.\beta(\bx_{i},\bw)}{\beta(\bx_{i},\bw).\Big(1+\beta(\bx_{i},\bw)\Big)}\bigg],
\end{equation}
where $\Psi(\cdot)$ is the digamma function defined as 
\begin{equation}
\Psi(\alpha)=\frac{d}{d\alpha} \log \left(\Gamma(\alpha)\right)=\frac{{\Gamma'(\alpha)}}{\Gamma(\alpha)}.
\end{equation}

\subsection{Inference}
\label{sec:inference}
Having the learned parameters of the network $(\bw^{*},\btheta^{*})$, for a test feature $\bx^{+}$, we use a MAP estimate to generate a set output as 
\begin{equation}
\calY^{*} 
= \arg\max_{\calY} p(\calY|\D,\bx^{+}),
 \end{equation}
where
\begin{eqnarray*}
	p(\calY|\D,\bx^{+}) & = & \int p(\calY|\bx^{+},\btheta,\bw)p(\btheta,\bw|\D)d\btheta d\bw
\end{eqnarray*}
and $p(\btheta,\bw|\D) = \delta(\btheta=\btheta^{*},\bw=\bw^{*}|\D)$. 
Since the unit of hypervolume $U$ in most practical application in unknown, to calculate the mode of the set distribution $p(\calY|\D,\bx^{+})$, we use the sequential inference as explained in~\cite{mahler2007statistical}. To this end, we first calculate the mode $m^*$ of the cardinality distribution 
\begin{equation}
m^{*} 
= \arg\max_{m}\quad p(m|\bw^*,\bx^{+}),
\end{equation}
where 
\begin{equation}
p(m|\bw^*,\bx^{+})=\NB\left(m;\alpha(\bw^*,\bx^{+}),\frac{1}{1+\beta(\bw^*,\bx^{+})}\right).
\end{equation}
Then, we calculate the mode of the joint distribution for the given cardinality $m^{*}$ as
\begin{equation}
\calY^{*}
= \arg\max_{\calY_{m^{*}}}\quad p(\{y_1,\cdots,y_{m^{*}}\}_{||}|\btheta^*,\bx^{+}).
\end{equation}
    To estimate the most likely set $\calY^{*}$ with cardinality $m^{*}$, we use the first CNN with the parameters $\btheta^*$ which predicts $p(y_1,\cdots,y_{M}|\bx^{+},\btheta^*)$, where $M$ is the maximum cardinality of the set, \ie $\{y_1,\cdots,y_{m^{*}}\}\subseteq\{y_1,\cdots,y_{M}\}\quad,\forall m^{*}$.
    Since the samples are \iid, the joint probability maximised when the probability of each element in the set is maximised. Therefore, the most likely set $\calY^{*}$ with cardinality $m^{*}$ is obtained by ordering the probabilities of the set elements $y_1,\cdots,y_{M}$ as the output of the first CNN and choosing $m^{*}$ elements with highest probability values. 
    
    % \fixmeh{We can say that this can be simply seen as the combination of cardinality estimation and any detector/classifier???}
    Note that the assumptions listed in \Sec~\ref{sec:deep-set-net0} are necessary to make both learning and inference computationally tractable and amenable to an elegant mathematical formulation.
    A major advantage of this approach is that we can use any state-of-the-art classifier/detector as the first CNN ($\btheta^*$) to further improve its performance. 
    Modifying any of the assumptions, \eg non-\iid set elements, leads to serious mathematical complexities~\cite{mahler2007statistical}, and are left for future work.

\section{Further Experimental Results}

Here, we provide additional arguments, evaluation plots and qualitative results that could not be included in the main paper.

\subsection{Object counting by regression}
\begin{table}[tb]
	%	\scriptsize
%	\small
	\caption{Loss comparison for cardinality estimation.}
	\label{tab:new-experiments}
	\vspace{-1.5em}
	\begin{center}
		\begin{tabular}{l|cc|cc}
			& \multicolumn{2}{c}{Mean card. error} & \multicolumn{2}{c}{F1 score}\\			
			\!\!\!\!  Loss & MOT16 & MS COCO & MOT16 & MS COCO \\
			\hline
			\!\!\!\! Regression  &$2.05$& $0.83$ & $60.16$ &$68.4$ \\
			\!\!\!\! Negative Binomial & $\textbf{1.94}$ & $\textbf{0.74}$ & $\textbf{61.86}$  & $\textbf{69.4}$ \\		
		\end{tabular}
	\end{center}
%	\vspace{-2.5em}
\end{table} 
Regressing for cardinality may seem an obvious choice, but is not trivial to derive mathematically and cannot be easily justified in our framework because it
\emph{a)} cannot be accommodated in a Bayesian set formulation to model uncertainty and 
\emph{b)} does not yield a discrete distribution. 
% since the cardinality is a discrete value, approximated with a continuous variable and a differentiable loss.
Nonetheless, we have conducted the experiment by replacing our loss with the regression loss while using the exact same architecture and setup as in \Sec 5.2 of the main text. Tab.~\ref{tab:new-experiments} shows the comparison results between our cardinality loss and regression loss on two datasets from two reported tasks of multi-class image classification (MS-COCO) and pedestrian detection (MOT16). As expected, directly regressing for cardinality yields slightly worse results both in terms of the cardinality prediction and \wrt the F1 score.
For completeness, \Tab~\ref{table:Card_err} also reports the mean absolute error and standard deviation for cardinality estimation using our loss on four  datasets.
\begin{table}[tb]
	% \small
	\caption{Mean absolute error and standard deviation for cardinality estimation on test sets.}
	\vspace{-1.5em}
		\begin{center}
		\begin{tabular}{l|cc|cc}
			& \multicolumn{2}{c}{Multi-label classification} & \multicolumn{2}{c}{Pedestrian detection}\\

			 Error& PASCAL VOC & MS COCO & Caltech & MOT16\\
			\hline\hline
			Mean  & $0.32$ & $0.74$ & $0.54$ & $1.94$\\
			\hline
			Std & $0.52$ & $0.86$ & $0.79$ & $1.96$ \\

		\end{tabular}
	\end{center}
	\label{table:Card_err}
		% \vspace{-1.5em}
\end{table}

\subsection{Pedestrian detection}

Here, we first discuss the challenges that we confronted to use our set formulation for this application. Then we provide some information about the datasets and their split used for this experiment. Finally, we show more quantitative and qualitative results on this experiment.

\myparagraph{Non-maximum suppression.}
In the main text, we argued that the non-maximum suppression (NMS) as a heuristic step makes the detection problem not as
straightforward as what is expressed in our set formulation
in \Eq~(\ref{eq:set_density0}).
In fact, a major nuisance in detection is not the NMS algorithm itself as a greedy solver, but rather its hand-crafted objective.
This process is traditionally formulated as maximising the joint distribution over pairs of samples, or equivalently as a quadratic binary program (QBP)
\begin{equation}Y^{*} 
= \arg\max_{Y} Y^TQY, \end{equation}
where
$Y\in\mathbb{B}^M$
is a binary vector, indicating which of the $M$ boxes to keep or to suppress.
The diagonal values of $Q$ are proportional to the detection scores while the pairwise (exclusion) terms in $Q$ are manually designed, \eg to correspond to the overlap ratios. 
The aforementioned QBP is NP-hard and cannot be solved globally in general. NMS is one greedy and efficient approach to solve the problem locally.
To enforce $m^*$, one could include a constraint into the QBP like
%The elegant way to formulate the detection problem in our set framework is to constraint this quadratic programming problem such that the output results necessarily contain $m$ boxes. Therefore, the problem can be defined as 
\begin{equation}
\begin{aligned}
Y^{*}  = &\arg\max_{Y}Y^TQY,\\ & s.t. \textbf{1}^TY = m^*.
\end{aligned} 
  \end{equation} 
However, this may lead to an infeasible problem for a fixed $Q$ with a predefined value of the threshold for an overlap ratio $T_O$. To this end, $Q$ should be designed such that the above problem can have a feasible solution.  Learning $Q$ is perhaps a more elegant approach, but is not part of this paper. To this end, for the current setup, one solution is to find a threshold that can make the above problem feasible. Therefore, we start from the default value of $T_O$, and adjust it step-wise until the number of boxes reaches $m^*$. In the case if the number of final boxes is larger than $m^*$, we pick $m^*$ boxes with the highest scores. To apply a solver, we experimented with the global QBP formulation using Gurobi for a small problem, but found NMS with an adaptive threshold to be the most efficient and effective approach.

\myparagraph{Caltech Pedestrians~\cite{Dollar:2012:PAMI}} is a de-facto standard benchmark for pedestrian detection. The dataset contains sequences captured from a vehicle driving through regular traffic in an urban environment and provides bounding box annotations of nearly $350,000$ pedestrians. The annotations also include detailed occlusion labels. The number of pedestrians per image varies between $0$ and $14$. However, more than $55\%$ of the images contain no people at all and around $30\%$ of the data includes one or two persons.    
We use the MS-CNN~\cite{Cai:2016:ECCV} network model and its parameters learned on the Caltech training set as $\btheta^*$ in \Eq~\eqref{eq:CNN_Eq0}. To learn the cardinality, we use $4250$ images provided as a training set, splitting it into training and validation ($80\%-20\%$), reaching a mean absolute error of $0.54$ (\cf~\Tab~\ref{table:Card_err}).
%Quantitative detection results are shown in \Tab~\ref{table:F1 score}. We achieve state-of-the art performance on this benchmark, improving on the currently best detector by applying our DeepSet network to estimate the number of persons in each image. 
%The improvement, however, is not as significant as in the case of multi-class image classification. We believe that this is mainly due to very limited variation of the number of pedestrians in the scene. For more than $85\%$ of the images containing $0$ to $2$ pedestrians. A single global threshold for a state-of-the art detection output appears to work almost as well as introducing the knowledge about the number of persons in the scene. Another reason for the relatively low improvement maybe the already remarkable performance of modern detectors on this dataset (\cf~\cite{Dollar:2012:PAMI}).
%\fixmeh{mean absolute error = $0.5393$ and std absolute error  = $0.7942$ on the test set}
%
%\fixmeh{Fo MS-CNN, max F1 score =  0.5161, mean F1 score = 0.1884. For deepset, max F1 score =  0.5215, mean F1 score = 0.2539. For deepset with GT, max F1 score =  0.5228, mean F1 score = 0.2503. }

\myparagraph{MOTCallenge 2016.}
%Additionally, we evaluate the MS-CNN detector and DeepSet in significantly more crowded and more challenging sequences from MOT16 dataset~\cite{Milan:2016:MOT16}. 
This benchmark is primarily targeted at multi-object tracking and is not yet commonly used for evaluating pedestrian detection. However, the variation in the number of pedestrians across the frames is relatively large (between $0$ and $32$) and is also distributed more uniformly, which makes correct cardinality estimation more important.
% Therefore, it will be more proper benchmark for representing the superior performance of our DeepSet compared to the Caltech dataset. 
Since the labels for the test set are not available, we use the provided training set of this benchmark consisting of $5316$ images from $7$ different sequences, and divide it into training, validation and test set with split ratios $60\%$, $15\%$ and $25\%$, respectively.  
We only learn the cardinality network $\bw^*$ on training set and we use the MS-CNN network model and its parameters learned on the KITTI dataset~\cite{Geiger:2012:CVPR} as $\btheta^*$ in \Eq~\eqref{eq:CNN_Eq0}. 
%The results are summarised in \Tab~\ref{table:F1 score} and show a more significant improvement over MS-CNN on this dataset compared to Caltech.
%%
%We believe that these results can be improved further by formulating a more principled way of choosing the $m^*$ predicted targets in the scene, as opposed to relying on the greedy NMS heuristic.

\begin{figure}[ht]
	\centering
	\includegraphics[width=.395\linewidth]{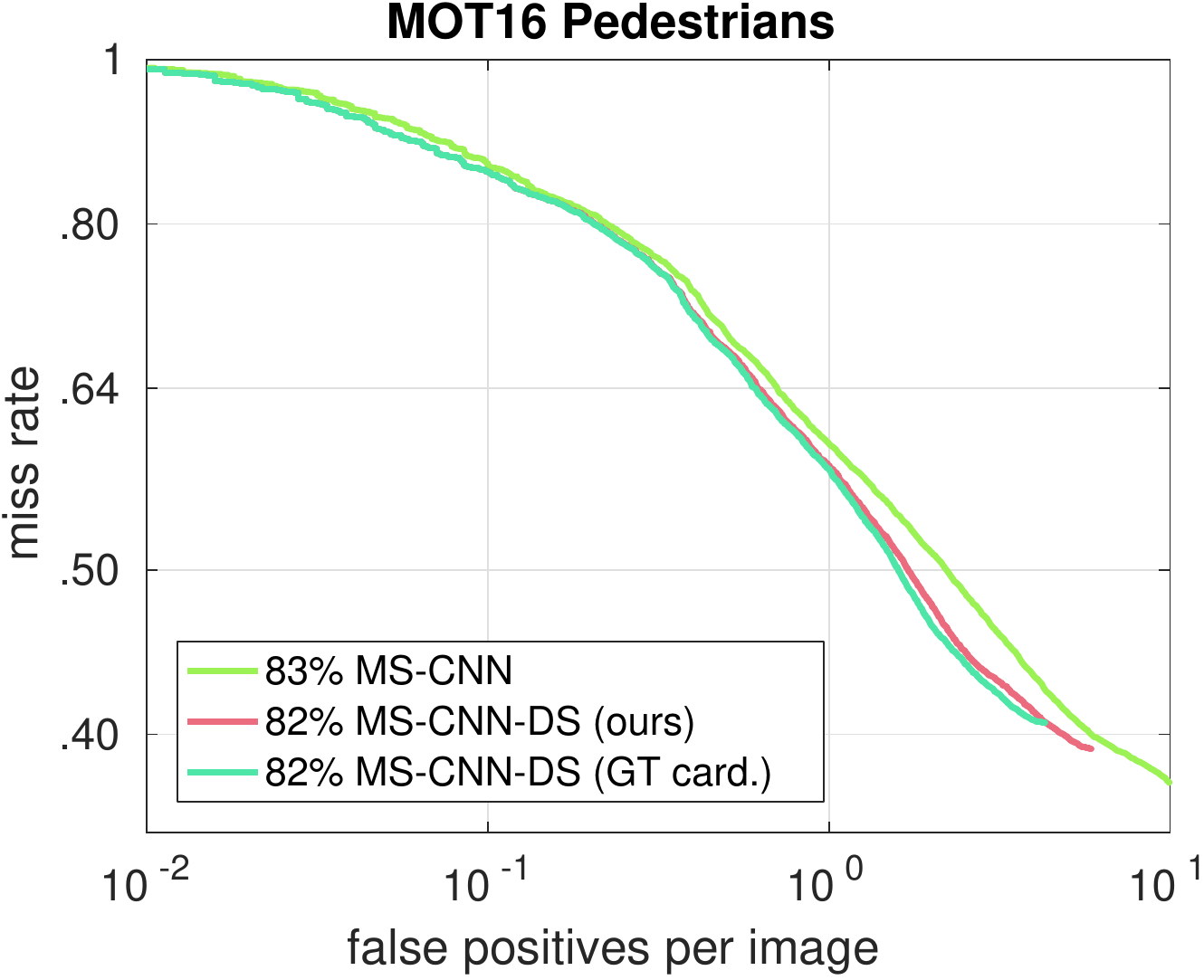}
	\includegraphics[width=.40\linewidth]{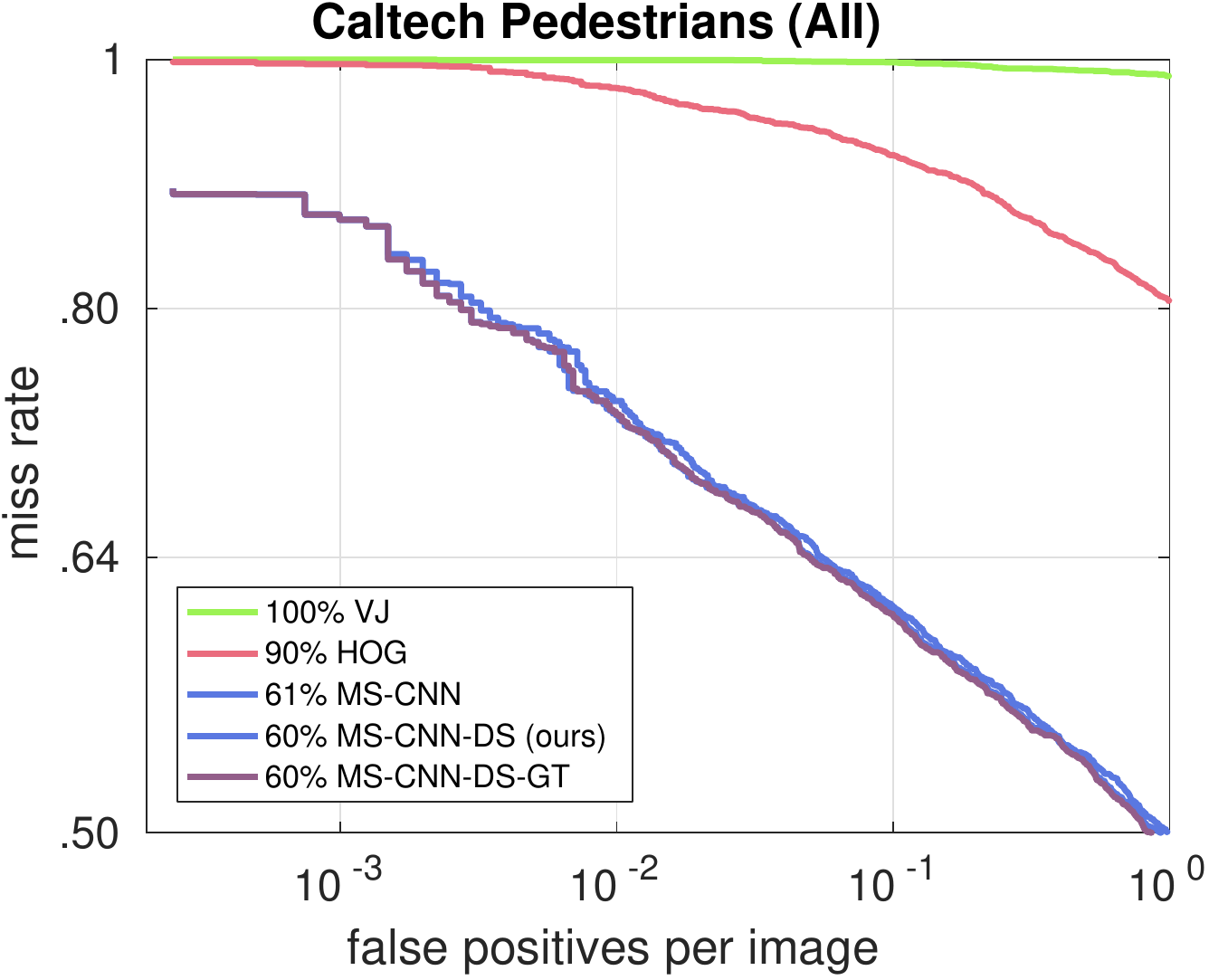}
	\caption{ROC curves on MOT16 and Caltech Pedestrians (experiment ``all''). The overall performance of a detector is measured by the log-average miss rate as proposed by Doll\'ar~\etal~\cite{Dollar:2012:PAMI}.}
	\label{fig:detection-plots}
\end{figure}

\begin{figure*}[t]
	\centering
	\includegraphics[width=1\linewidth]{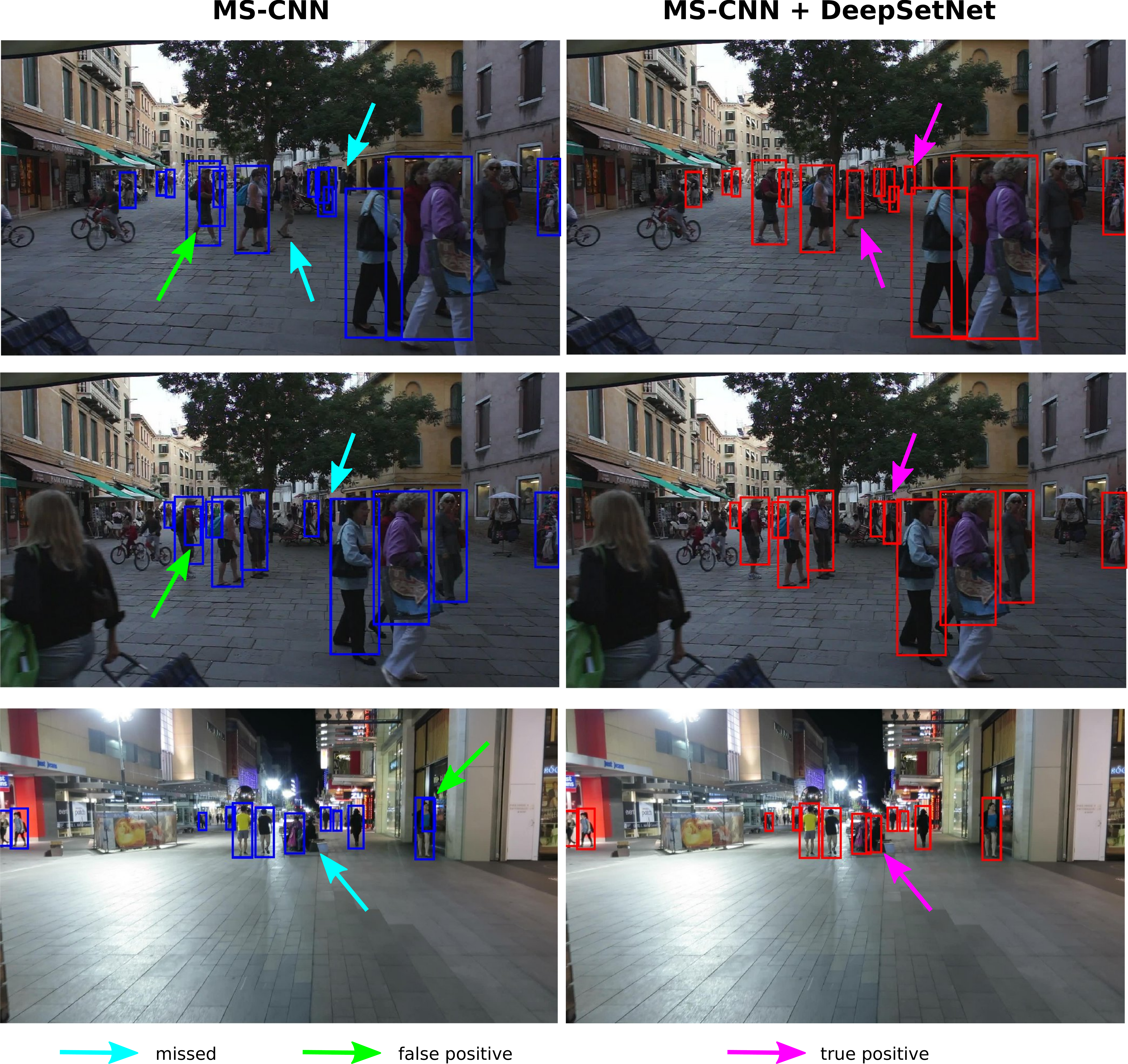}
	\caption{More examples illustrating results of pedestrian detection generated using the state-of-the-art detector MS-CNN~\cite{Cai:2016:ECCV} (in blue, left) and our MS-CNN + DeepSetNet (right).
		To generate the MS-CNN results, we display the top $m^*$ boxes after applying non-maximum suppression. Arrows indicate typical failures introduced by a suboptimal NMS threshold, which are corrected when considering the predicted number of persons for each image.
	} 
	\label{fig:Results3}
\end{figure*}

\myparagraph{Additional Results.} ROC curves on two detection datasets are shown in Fig.~\ref{fig:detection-plots}.
Qualitative results of pedestrian detection are shown in Figure~\ref{fig:Results3}.

%-------------------------------------------------------------------------
%\subsection{Multi-label Classification}

\subsection{Multi-class image classification.}
\begin{figure*}[t]
	\centering
	\includegraphics[width=.95\linewidth]{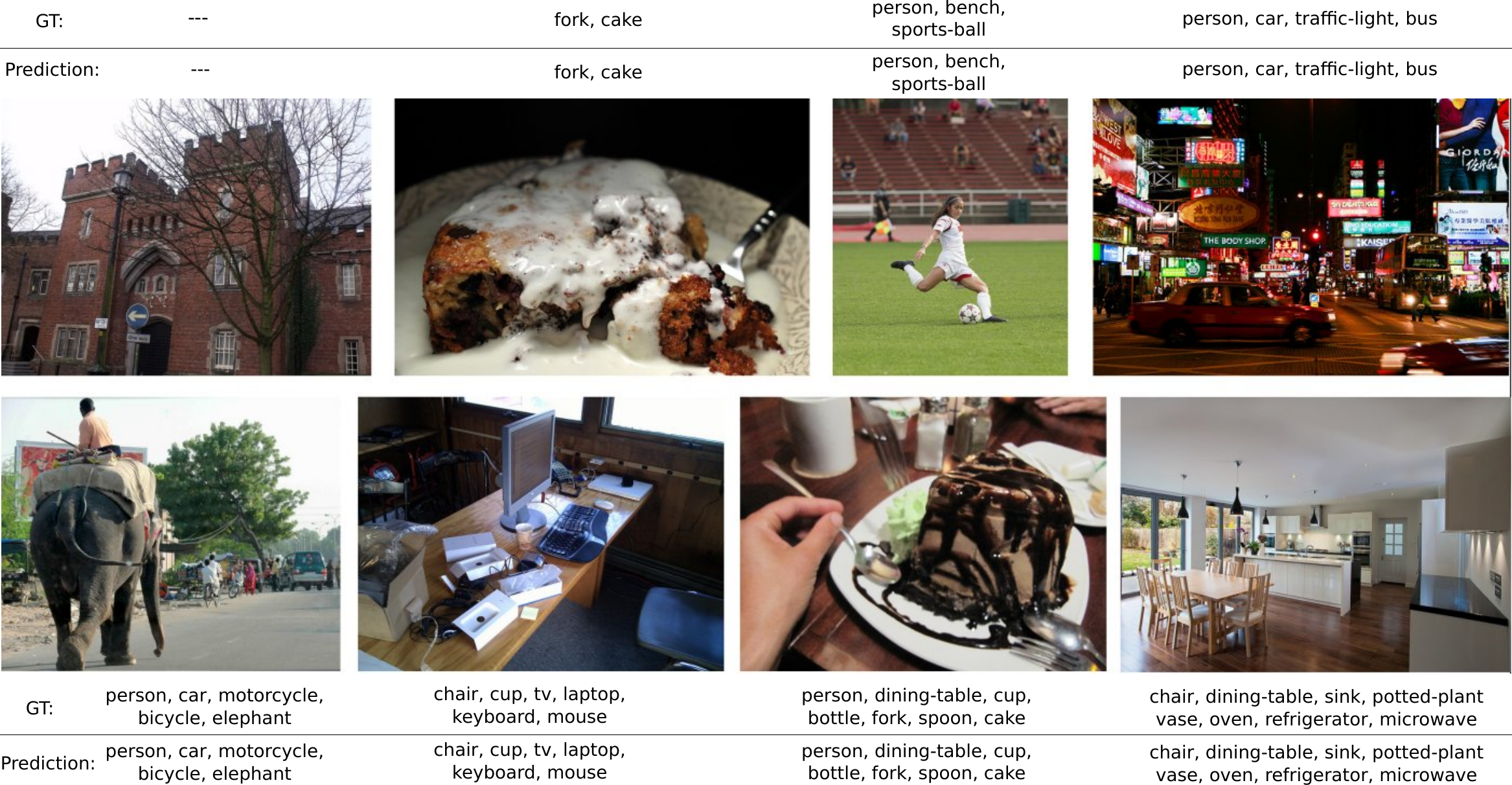}
	\caption{Further examples showing a perfect prediction \wrt both the number of tags and the labels themselves using our Deep Set Network.} 
	\label{fig:Results10}
\end{figure*}

\begin{figure*}[t]
	\centering
	\includegraphics[width=.95\linewidth]{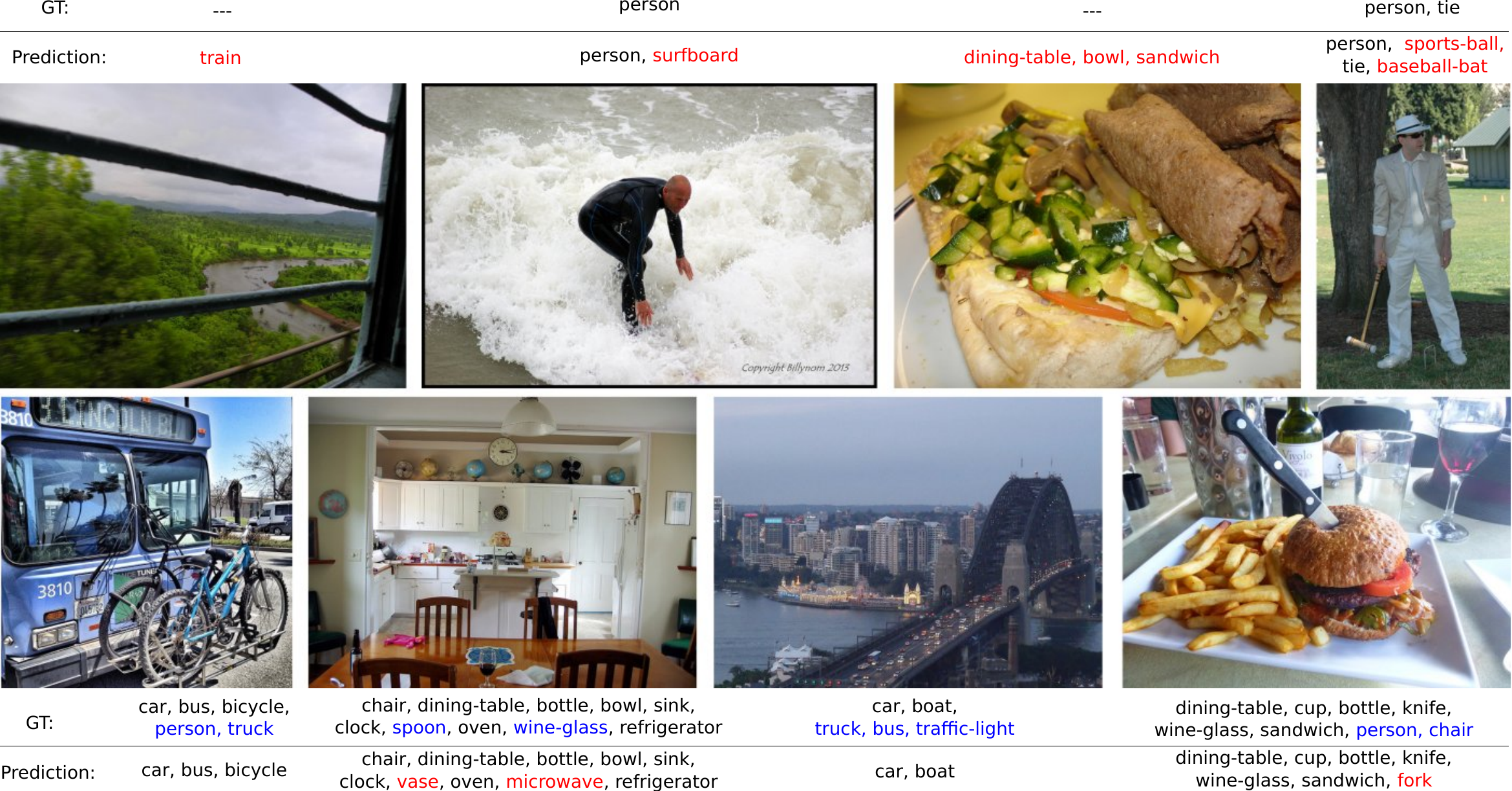}
	\caption{Additional examples illustrating interesting failure cases. {\textcolor{blue}{False negatives}} and {\textcolor{red}{false positives}} are highlighted in blue and red, respectively. Note that in most examples, the mismatch between the ground truth and our prediction is due to the ambiguity or artifacts in the annotations. Two such examples are shown in the top row, where a train (window) and the surfboard are not annotated, probably because these are hardly visible in the image. Nevertheless, our network can predict the objects.
		The two bottom rows show real failure cases of our method. Note, however, that these include extremely challenging examples, where even for a human, it is fairly difficult to spot a traffic light in the aerial image or the person and the chair in the image on the bottom right.
	} 
	\label{fig:Results20}
\end{figure*}

Figure~\ref{fig:Results10} shows more results for successful image tagging. Figure~\ref{fig:Results20} points to some interesting failures and erroneous predictions. 


\end{document}